\documentclass{article}

\PassOptionsToPackage{numbers, compress}{natbib}

\usepackage[preprint]{neurips_2023}

\usepackage[utf8]{inputenc} 
\usepackage[T1]{fontenc}    
\usepackage{hyperref}       
\usepackage{url}            
\usepackage{booktabs}       
\usepackage{amsfonts}       
\usepackage{nicefrac}       
\usepackage{microtype}      
\usepackage{xcolor}         
\usepackage{tikz}
\usepackage{amsmath}
\usepackage{xspace}
\usepackage{listings}
\usepackage{enumitem}
\usepackage{mdframed}
\usepackage{multirow}
\usepackage{tabularx}  
\usepackage{subcaption}
\usepackage{fancyvrb}

\lstdefinelanguage{yaml}{  
  keywords={true,false,null,y,n},  
  keywordstyle=\color{darkgray}\bfseries,  
  basicstyle=\small\ttfamily,  
  comment=[l]{\#},  
  commentstyle=\color{purple}\ttfamily,  
  stringstyle=\color{red}\ttfamily,  
  morestring=[b]',  
  morestring=[b]"  
}
\usepackage{listings}  
\usepackage{xcolor}  
  
\newcommand{\method}{TaskWeaver\xspace}

\title{\includegraphics[width=1.2cm]{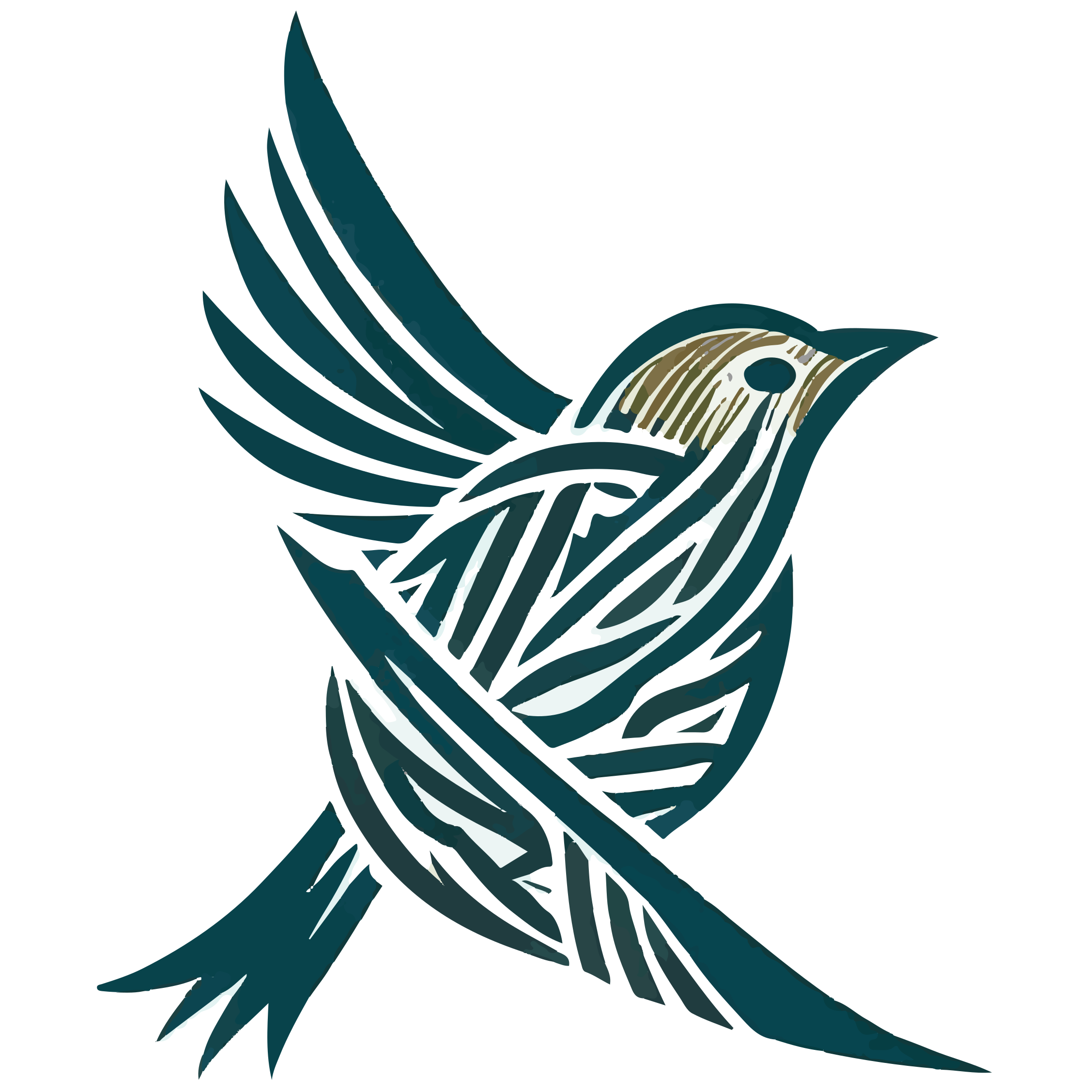}
\method: A Code-First Agent Framework}

\author{
Bo Qiao\thanks{Equal Contribution} \And Liqun Li\footnotemark[1] \And Xu Zhang\footnotemark[1] \And Shilin He\footnotemark[1]\And  Yu Kang  \And Chaoyun Zhang \And Fangkai Yang \And Hang Dong \And Jue Zhang \And Lu Wang \And Minghua Ma \And Pu Zhao \And Si Qin 
 \And  Xiaoting Qin  \And Chao Du \And Yong Xu  \And Qingwei Lin \And Saravan Rajmohan  \And Dongmei Zhang\\ 
 Microsoft\\
\texttt{taskweaver@microsoft.com}\\
}

\begin{document}

\maketitle

\begin{abstract}
Large Language Models (LLMs) have shown impressive abilities in natural language understanding and generation, leading to their widespread use in applications such as chatbots and virtual assistants. However, existing LLM frameworks face limitations in handling domain-specific data analytics tasks with rich data structures. Moreover, they struggle with flexibility to meet diverse user requirements.
To address these issues, TaskWeaver is proposed as a code-first framework for building LLM-powered autonomous agents. It converts user requests into executable code and treats user-defined plugins as callable functions. TaskWeaver provides support for rich data structures, flexible plugin usage, and dynamic plugin selection, and leverages LLM coding capabilities for complex logic. It also incorporates domain-specific knowledge through examples and ensures the secure execution of generated code. TaskWeaver offers a powerful and flexible framework for creating intelligent conversational agents that can handle complex tasks and adapt to domain-specific scenarios. The code is open-sourced at \href{https://github.com/microsoft/TaskWeaver/}{https://github.com/microsoft/TaskWeaver/}.
\end{abstract}

\section{Introduction}  
   
Large Language Models (LLMs), such as GPT \cite{radford2018improving, brown2020language}, Claude \cite{claude}, Palm \cite{anil2023palm}, Gemini \cite{geminiteam2024gemini}, Llama \cite{touvron2023llama}, and Mixtral \cite{jiang2024mixtral} have demonstrated remarkable capabilities in natural language understanding and generation. These models have been widely used in various applications, including chatbots, virtual assistants, and content-generation systems. There is a growing potential for LLMs to revolutionize the way humans interact with machines, providing a more natural and intuitive experience.  

An agent, specifically those that utilize Large Language Models (LLMs) or other AI technologies, is regarded as an autonomous entity that possesses the ability to plan tasks, observe its surroundings, and execute appropriate actions accordingly~\cite{wang2023survey, xi2023rise}.
Several existing frameworks, including Langchain \cite{langchain}, Semantic Kernel \cite{semantickernel}, Transformers Agent \cite{transformeragents}, Agents \cite{zhou2023agents}, AutoGen \cite{wu2023autogen}, XAgent \cite{xagent2023}, and JARVIS \cite{jarvis}, have endeavored to utilize LLMs for general-purpose task-oriented conversations.
Other frameworks such as Open Interpreter \cite{openinterpreter}, Cradle \cite{weihao2024cradle}, and UFO \cite{ufo} focus on more specialized tasks such as controlling the computer system. 
A recent work \cite{hong2024data} introduces a Data Interpreter for the multi-agent framework MetaGPT \cite{hong2023metagpt} to enhance its power in data analytics. 
These frameworks enable users to interact with LLM-powered agents by issuing natural language requests and receiving responses in return. Nevertheless, these frameworks possess limitations that constrain their efficacy in handling domain-specific scenarios and data analytics tasks.
   
One major limitation is that most existing frameworks lack native support to handle rich data structures. LLM-powered agents often need to work with complex data structures, such as nested lists, dictionaries, or data frames, for data analytics applications and many other business scenarios. However, many of existing frameworks struggle to handle these structures efficiently, particularly when it comes to transferring information between chat rounds or across different plugins. In such cases, these frameworks either persist data to disk or encode complex structures as strings or JSON objects in the prompts. Although these approaches are functional, they can lead to impracticality and increased error rates, particularly when dealing with large datasets.

Another limitation of existing approaches is the lack of configuration for incorporating domain knowledge. While these frameworks provide tools and examples for prompt engineering, they fail to offer a systematic way to embed domain-specific knowledge into the planning and code-generation process. Consequently, the limitation makes it challenging to control the planning and code generation process  in accordance with specific domain requirements.

Another issue encountered in many existing frameworks is their inflexibility, which hinders the ability to meet the diverse requirements of users. Although plugins can address common needs, they may fall short when it comes to handling ad-hoc queries. Writing a separate plugin for each ad-hoc query is impractical. In these situations, it becomes necessary for the agent to be able to write custom code to execute the user's query. Therefore, there is a need for a solution that seamlessly integrates plugin execution with custom code execution to address this issue.

To address these limitations, we propose \method, a code-first framework for building LLM-powered autonomous agents. The standout feature of \method is its ability to convert each user request into executable code, treating user-defined plugins as callable functions. \method overcomes the limitations of existing frameworks by providing support for rich data structures, flexible plugin usage, and dynamic plugin selection. It leverages the coding capability of LLMs to implement complex logic and incorporates domain-specific knowledge through examples. Additionally, \method has made considerable efforts towards the secure execution of generated code and provides an easy-to-use interface for developers. 

In this paper, we present the design and implementation of \method, along with several case studies that demonstrate its effectiveness in handling various tasks. Overall, \method provides a powerful and flexible framework for building intelligent conversational agents that can handle complex tasks and adapt to domain-specific scenarios.

\section{Motivation and Requirements}\label{sec:motivation}

To illustrate the \method approach, let's consider a real-world use case – conducting anomaly detection on time series data stored in an SQL database. Our goal is to apply a specialized anomaly detection algorithm to this data, which requires two input columns: one of type timestamp and the other of type float. The expected conversation between the user and the AI assistant proceeds as follows:

\begin{figure}[htbp]
    \centering
    \includegraphics[width=3in]{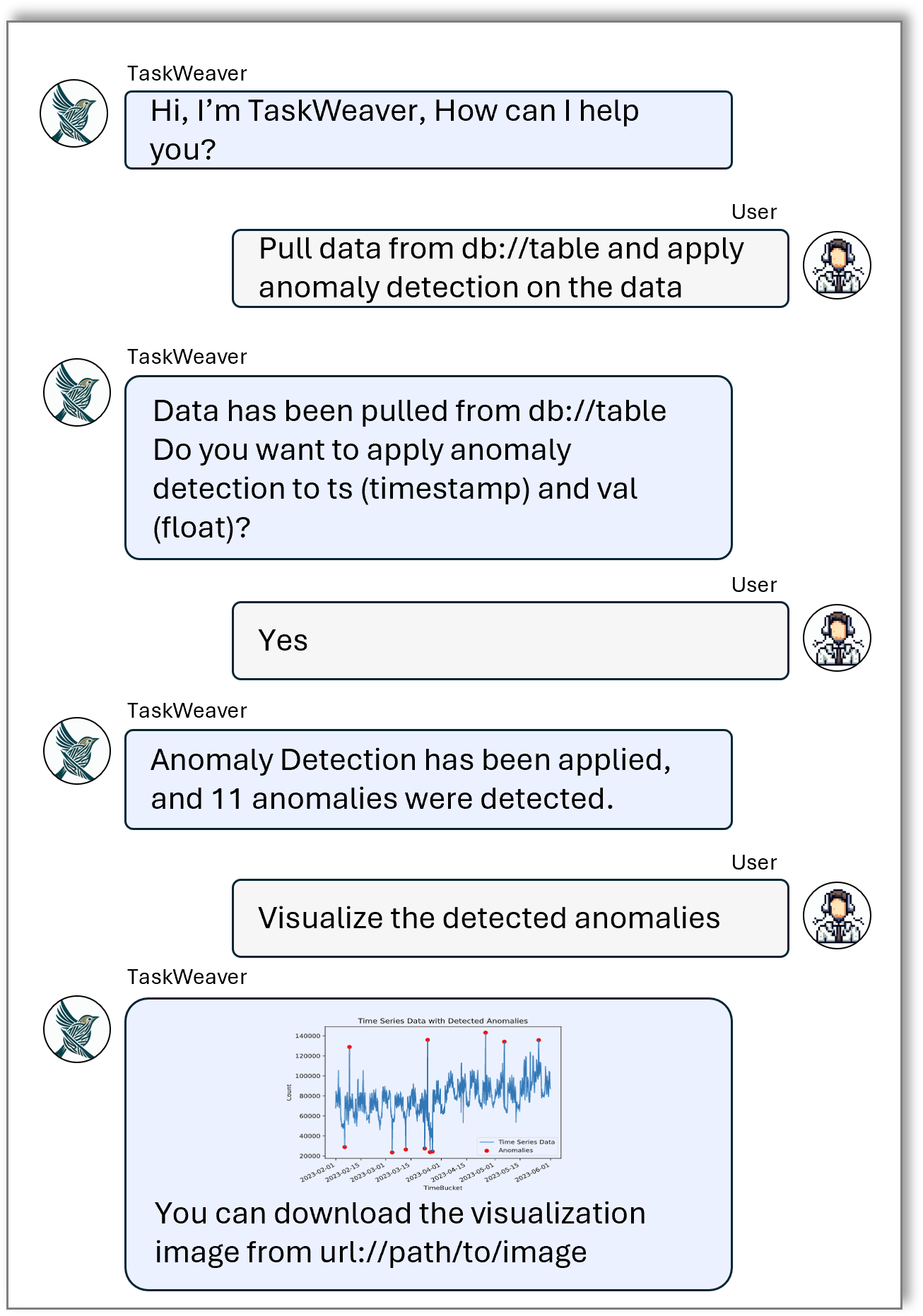}
    \caption{Chat between the user and the assistant.}
    \label{fig:chat}
\end{figure}

The example demonstrates common actions in a data analysis task, which include obtaining data, performing analysis on the data, and visualizing the results. By examining the steps in this use case, we can identify key requirements imposed by this example:  
   
\begin{itemize}[leftmargin=*]
    \item Plugin: \method must support invoking custom plugins. In this example, plugins are needed to pull data from the database and implement the specialized anomaly detection algorithm.  
    \item Rich data structure: \method must be capable of handling data in complex structures, such as pandas DataFrame, to perform advanced data processing actions. Data in rich structure should be able to transfer easily from one plugin to another. 
    \item Stateful execution: \method engages in iterative interactions with the user, processing user inputs and executing tasks accordingly. The execution state is preserved throughout the entire conversation session across multiple chat rounds.
    \item Reasoning and action (React): \method is unaware of the data schema stored in the database prior to reading it. To generate the anomaly detection code, \method must first inspect the data schema and then input the corresponding column names into the anomaly detection algorithm.
    \item Response in natural language: \method consistently responds to the user in human-readable natural language. Generally, the anomaly detection algorithm returns a DataFrame, but \method needs to provide a summary of the execution result, such as ``\textit{11 anomalies were detected}''.  
    \item Code generation: \method must generate code to accommodate ad-hoc user demands, which are not covered by the pre-defined plugins. In the example provided, \method generates code to visualize the detected anomalies.  
    \item Incorporating domain knowledge: \method should provide a systematic way to incorporate domain-specific knowledge.
    It would help LLMs make better planning and accurate tool calls, which in turn produces reliable results, particularly in complex domains.
    \item Persisting artifact: \method should offer a means of saving results, such as DataFrames or images, to persistent storage. Users can download the artifacts via the provided links. 
\end{itemize}

We have so far summarized the requirements from our motivating example. In the following section, we are going to describe the design of \method.

\begin{figure}
    \centering
    \includegraphics[width=5in]{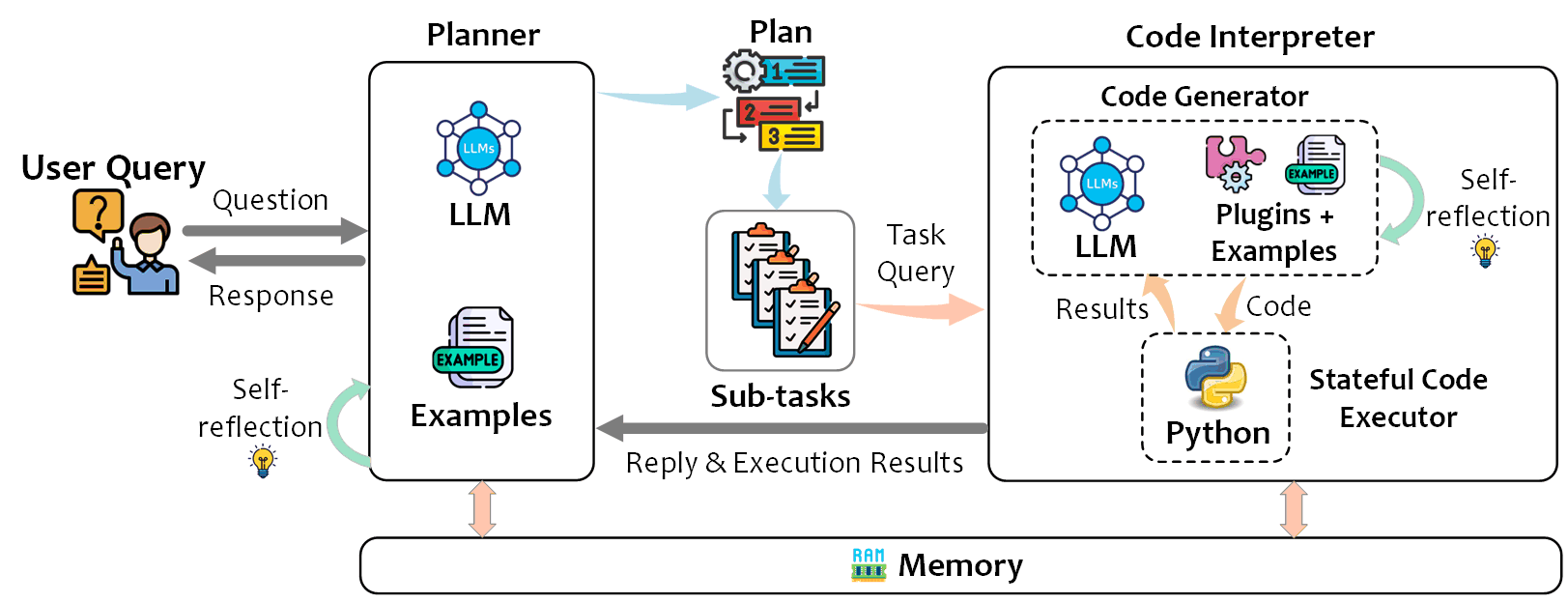}
    \caption{The overview of \method}
    \label{fig:overview}
\end{figure}

\section{An Overview of \method}\label{sec:overview}
Fig. \ref{fig:overview} presents an overview of \method, which consists of two key roles: the \textit{Planner} and the \textit{Code Interpreter} (CI). The CI consists of a \textit{Code Generator} (CG), and a \textit{Code Executor} (CE). The Planner serves as the system's entry point and interacts with the user. Its responsibilities include: (1) planning – breaking down the user's request into subtasks and managing the execution process with self-reflection; and (2) responding – transforming the execution result into a human-readable response for the user. 

The CI is responsible to generate code snippets for any given task and run them to obtain the execution result. Specifically, the CG generates code for a given subtask from the Planner, considering existing plugins to enable the generated code to incorporate function calls for specific tasks. The examples within the CG guide it, particularly for domain-specific tasks unfamiliar to the LLM. Lastly, the CE is responsible for executing the generated code and maintaining the execution state throughout the entire session.

\begin{figure}
    \centering
    \includegraphics[width=5.5in]{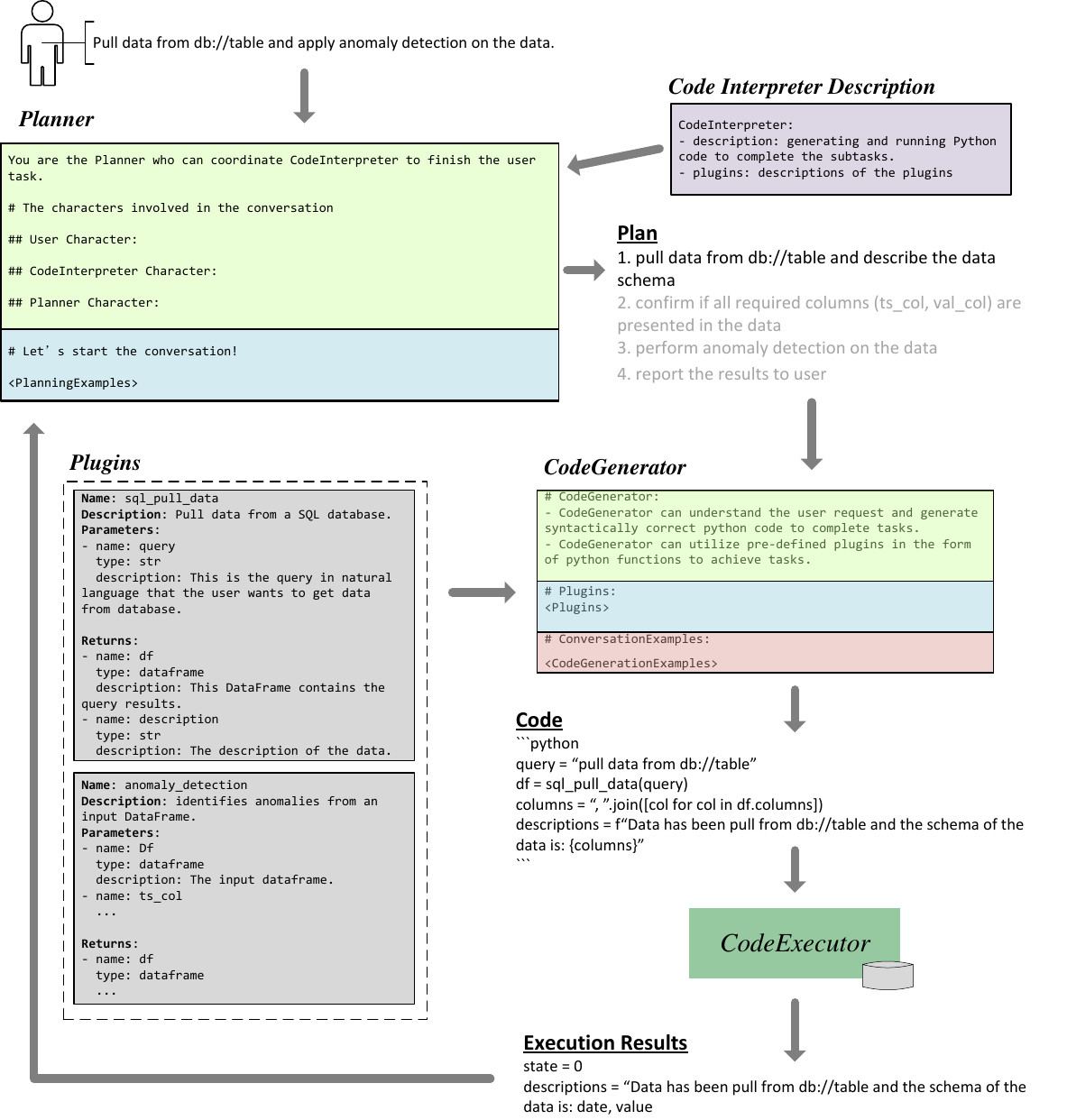}
    \caption{Workflow of \method}
    \label{fig:workflow}
\end{figure}

A centralized Memory module maintains the chat history of the current conversation session between the user and \method's internal roles (i.e., the Planner and Code Interpreter). In addition, the Memory also stores useful information for each role, such as the step-wise plans of the Planner, the thoughts and the generated code snippets of the Code Interpreter, etc. These data are usually referred to as short-term memory as they are only relevant to the current conversation. \method also has its long-term memory that can be shared across conversations, which will be detailed in the following sections.

Recall that we have motivated our design with an example of pulling data from a database and applying a custom anomaly detection algorithm to the data. We now explain how a sub-task of it is accomplished in \method. The workflow is illustrated in Fig. \ref{fig:workflow}. The prompts shown in Fig. \ref{fig:workflow} is simplified and only for illustration purpose due to the space limitation. The actual ones are much more complicated. 

The initial step involves the Planner taking the user query, Code Interpreter (CI) description, and, if provided, planning examples to generate a plan. The CI description outlines its code generation and execution capabilities. To enhance the Planner's effectiveness in task planning, the CI description lists the available plugins that is callable by CI. The output of the Planner is a step-by-step plan, according to which the Planner phrases the queries and communicates with the CI. The first step (highlighted in Fig. \ref{fig:workflow}) consists of pulling data from the database and describing the data schema.

The Code Generator (CG) prompt delineates its profile and competencies, providing comprehensive definitions of all the relevant plugins. This includes the function name, its description, the arguments it accepts, and what it returns. Additionally, code generation examples may be incorporated into the prompt to steer the code generation process. The output from the CG is a code snippet that executes the \textit{sql\_pull\_data} plugin, retrieves the data into a DataFrame, and describes the data schema.

The Code Executor (CE)'s execution result is sent back to the Planner to determine the next step in the plan. In practice, the Planner may modify its original plan if the outcome differs from expectations. In our example, the execution result reveals two columns, namely \textit{date} and \textit{value}, in the DataFrame. For the next step, the Planner can either confirm with the user if these columns correspond to the two input parameters \textit{ts\_col} and \textit{val\_col} of the \textit{anomaly\_detection} plugin, or directly proceed to the third step, as it is reasonable to assume that \textit{date} and \textit{value} likely represent \textit{ts\_col} and \textit{val\_col}, respectively. Regardless, \method must first retrieve the data and understand its schema before making a decision for the second step, which involves a self-reflection process.

As shown in this example, \method incorporates a two-layer planning process during the handling of user requests. The first layer consists of the Planner generating a high-level plan outlining the steps required to fulfill the request. Subsequently, in each round, the CI must devise a plan, in terms of chain-of-thought and generated code, to execute the specified step.

\subsection{Concepts}
We introduce some important concepts in the \method system.

\paragraph{Session} 
A new \textit{session} is established once the user submits their initial request to the \method. This session is terminated if the user intentionally resets the conversation or when a predetermined expiration time is reached following the last interaction. There could be multiple concurrent sessions running in parallel serving different users. 

\paragraph{Round}
A \textit{round} commences with \method receiving a user request and concludes upon responding to the user. Subsequently, \method awaits the user's next input. Typically, a session consists of numerous rounds. Responses to the user may either involve a message after completing a specific task or \method requesting further input to successfully carry out the task.

\paragraph{Post}
Within a round, there can be several messages exchanged between the Planner and CI, each of which is referred to as a \textit{post}. A post contains the text message, such as the request from the Planner to CI for code generation, and may also include other relevant information like the generated code and execution logs, referred to as \textit{attachments}. Each attachment has a type and its content. 

\begin{figure}
    \centering
    \includegraphics[width=5in]{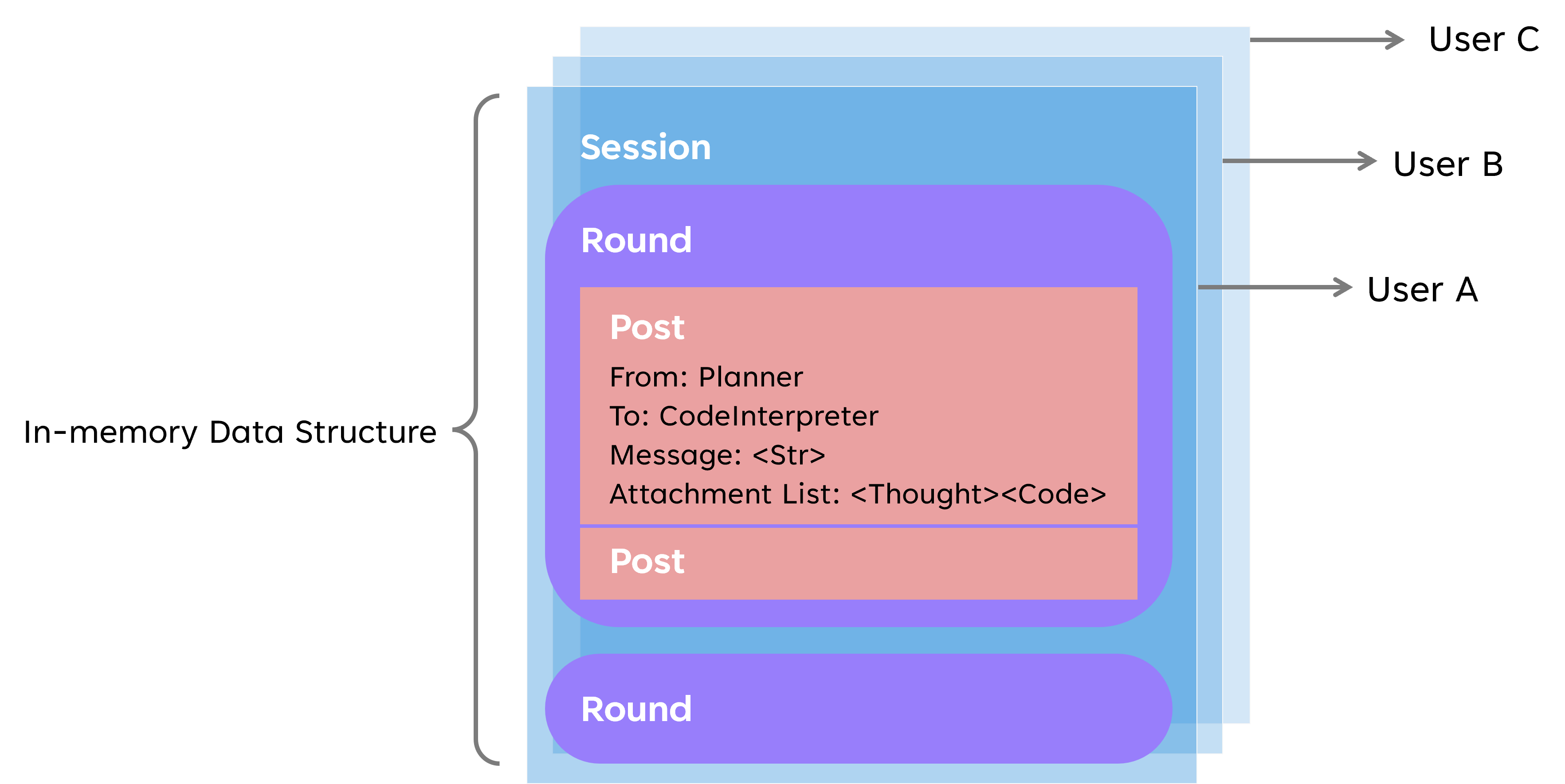}
    \caption{Concurrent sessions serving different users in parallel.}
    \label{fig:sessions}
\end{figure}

The in-memory data structure of Sessions, Rounds, and Posts is illustrated in Fig. \ref{fig:sessions}. 

\paragraph{Example}
It is common practice to use in-context learning for guiding LLMs in content generation, which involves adding examples to the prompt to demonstrate the `input's and their correspondent `output's. In \method, we have pre-defined two types of examples for the Planner and the CG, respectively. The examples for the Planner contain the user's request and desired steps of subtasks, while each example of the CG has the task description as input and generated code snippet as output. All examples follow the Rounds and Posts format and save in YAML files. 

\paragraph{Plugin}
Plugins in \method represent user-defined functions that can be invoked in the generated code snippets. Adding new plugins is the major way of extending \method's capabilities, especially in handling complex tasks that require domain-specific knowledge. In \method, Plugins are attached to the session, and therefore, different sessions can have different sets of plugins. 

\paragraph{Role}
In \method, a \textit{role} is conceptualized as an object instance capable of participating in a conversation by implementing a `reply' interface. Within this framework, we identify the Planner and the CI as two pre-defined roles integral to \method's functionality. Notably, \method is designed with flexibility in mind, allowing for the incorporation of additional roles. These roles are dynamically identified and integrated into the system during runtime. A distinctive attribute of the Planner role is its ability to communicate directly with the user, setting it apart from other roles such as the CI, which are constrained to interact solely with the Planner. This hierarchical communication model ensures structured and organized interactions within \method's architecture.


\section{Design Considerations}
In this section, we discuss the design considerations.

\subsection{Code-First Analysis Experience}

Python has emerged as the de-facto language for data analysis, and \method leverages this popularity by converting user requests into Python programs that run on dedicated processes. Users can issue natural language requests to manipulate the state of the Python process with \method, which utilizes popular libraries like numpy, pandas, sklearn, etc. Unlike other frameworks that rely on text or file-based expressions, \method utilizes native data structures such as pandas DataFrame. This makes it easy to perform tasks such as pulling data from a database, running machine learning algorithms (e.g., anomaly detection, classification or clustering), summarizing results, and visualizing analysis outcomes.

\subsection{Restricted Code Generation}
The \method approach involves a process of code verification (before code execution) to ensure that the generated code conforms to pre-defined rules and does not violate any safety requirements. This is achieved through the use of configurable rules that can be customized to suit different scenarios. By doing so, users can ensure that the code does not import any forbidden packages or invoke any unsafe functions. 
\method also supports a dedicated plugin-only mode, leveraging the function calling capability of LLMs, where only the plugin functions can be called, and all other generated code is denied.  

\subsection{Stateful Code Execution}

When users make ad-hoc requests for data analysis, it often involves multiple iterations. As a result, \method needs to maintain the state of code execution throughout the entire session. This is similar to programming in Python using a Jupyter Notebook, where users type code snippets in a sequence of cells and the program's internal state progresses sequentially. However, in \method, users use natural language instead of a programming language. \method converts each user request into one or more code snippets in each round, depending on the specific plan. 

\subsection{Intelligent Plan Decomposition}\label{subsec:intelli_plan}
The Planner decomposes users' requests into sub-tasks that are simplified and executable by the CI. We identified three types of dependencies between different sub-tasks: sequential, interactive, and none. Sub-task \textit{B} is sequentially dependent on sub-task \textit{A} means that \textit{B} must run after \textit{A} is done. Interactive dependency is the same as sequential in the execution order. More than that, it means there is a need for intervention by a human or a LLM between the two sub-tasks. For example, the user instructs \method to read a file and follow the instructions in its content. The Planner split the request into 2 steps: (1) read the file content; (2) follow the instructions. The LLM needs to complete the first step to read the content before carrying out the second step. The third type of dependency is none which means there is no dependency there, and they can be conducted in parallel. According to this definition, two steps where one sequentially depends on the other could be merged into one step for code generation because no intervention is required. In \method, the Planner involves two-phase planning, where we enforce it to reconsider the generated plan of sub-tasks to merge sub-tasks with sequential dependency. This can prevent the model from splitting the request into too fine-grained steps leading to a prolonged execution process and incurring unnecessary costs by calling the LLM many times. 

\subsection{Self-Reflection}
\method is designed with the capacity to rectify errors throughout the planning and code generation stages. When it detects that the outcomes of the preceding steps diverge from the anticipated results, the Planner possesses the capability to reassess and modify its plan, exploring alternative approaches. Furthermore, the CI is equipped to evaluate the results of code execution. Should the code execution process encounter an exception or the code fails to pass the verification, the CI can initiate a re-generation of the code, thereby attempting to correct the code. This iterative process ensures that \method maintains a robust and adaptable approach to task execution, increasing the reliability and efficiency of the framework.

\subsection{Scalable Plugin Usage}

In \method, plugins are specialized Python functions used to handle tasks that are either too complex or require specific domain knowledge, thereby eliminating the number of plugins needed since \method can already handle general Python code generation.   
Furthermore, \method features dynamic plugin selection. This means that after a user request is received, only the plugins that are relevant to that request are selected from a pool of available plugins. This approach ensures that \method uses the most appropriate tools for the task at hand without overloading the prompt with unnecessary functions.

\subsection{Incorporating Domain Knowledge}

One way to incorporate domain knowledge is by defining custom plugins, as discussed previously. However, for domain-specific tasks, it can be challenging for the LLM to generate the correct code to call the plugins or to make a good plan. To address this, we have introduced an interface in \method that allows users to guide the system for such difficult tasks. \method enables users to configure examples to teach the LLM how to respond to certain requests. For instance, a conversation history containing step-by-step thoughts and request/response sections can be used as an example. There are two types of examples in \method: one is used for planning and the other for code generation. By using examples, \method can incorporate domain-specific knowledge and improve the LLM's ability to generate accurate plans and code for difficult tasks. 

\subsection{Security and Safety}

Executing freely generated code can introduce security risks. For example, a malicious user may ask \method to open a file containing security keys, delete system files, or terminate a process. A typical way to mitigate the security risks is to include certain statements in the prompts to the LLM, which is implemented in \method.
To further prevent these malicious behaviors, \method allocates a separate worker process running inside a session-associated docker container. 
This architectural choice ensures that each worker process operates independently, isolated from processes of the host OS as well as other sessions, thus effectively preventing malicious behaviors.

\subsection{Usability}

Existing LLM frameworks such as Langchain \cite{langchain} make it easy to build proof-of-concept demos. However, building a reliable system with these frameworks can be time-consuming due to the large number of components they provide (like a large box of Lego bricks). One of the main goals of \method is to make it easy to use. Typically, users only need to customize the plugins to get started. For more difficult tasks, users can customize the examples for code generation and planning. To aid users, \method includes various tools to help, such as a tool to convert Python functions into plugins and a tool to save existing conversations as examples. These tools make it easier for users to customize the system. 
Another key feature of \method is its support for multi-tenant. The system implements a session manager to isolate different user sessions, making it easy to serve multiple users as a service.

\subsection{Cost Effectiveness}

The cost of calling LLMs can be significant, and \method addresses this issue by letting different roles be configured with different LLM models. For instance, GPT 3.5 is much cheaper compared to GPT 4. Therefore, for simpler tasks, we may use a cheaper model to reduce the overall cost. 

\section{\method in Detail}

We are going to explain each module in one section accordingly.

\subsection{Planner}
The Planner serves as a crucial component in the \method system, where it decomposes requests from users into several sub-tasks and orchestrates capabilities within \method to complete the task and report back to the users. From a high-level perspective, the Planner functions as the entry point and controller, managing the entire system.
As illustrated in Figure~\ref{fig:overview}, the Planner communicates bidirectionally with each component, sending queries and receiving responses.

As shown in Fig. \ref{fig:workflow}, a typical working routine of the Planner is as follows: The Planner first receives a query from users and then decides to decompose it into multiple sub-tasks. These sub-tasks essentially form a ``Initial Plan'', which is generated based on the knowledge of LLMs or enhanced by domain-specific ``Examples''. After drafting the initial plan, the Planner is requested to refine this initial plan by considering the dependencies among the sub-tasks as discussed in Sec. \ref{subsec:intelli_plan} in a chain-of-thought manner. The Planner may merge multiple sub-tasks into one in its ``Final Plan''. Table \ref{tab:plan} shows two examples of the Planner's initial plan and final plan w.r.t the User's request. In the first example, the initial plan has 4 steps while the first 3 steps are merged into one in the final plan. In the second example, the final plan is the same as the initial plan because the Planner needs to read the file content and then understand what would be the next step.

\begin{table}[ht]  
\begin{tabularx}{\columnwidth}{X|X}  
\hline  
\multicolumn{2}{l}{\textbf{User request: Load data from a.csv, show the column names, and count the number of rows}} \\ \hline  
Initial Plan & Final Plan \\ \hline  
\begin{enumerate}[leftmargin=*]  
  \item load a.csv  
  \item extract and display the column names <sequentially depends on 1>  
  \item count the number of rows <sequentially depends on 1>  
  \item report the information to the user <interactively depends on 2,3>  
\end{enumerate}  
&   
\begin{enumerate}[leftmargin=*]
  \item load a.csv, extract the column names, and count the number of rows  
  \item report the information to the user  
\end{enumerate} \\ \hline  \hline
\multicolumn{2}{l}{\textbf{User request: Read file manual.txt and follow the instructions in it}} \\ \hline  
Initial Plan & Final Plan \\ \hline  
\begin{enumerate}[leftmargin=*] 
  \item read manual.txt and show its content  
  \item follow the instructions according to the file content <interactively depends on 1>
  \item report the result to the user <interactively depends on 2>  
\end{enumerate}  
&   
\begin{enumerate}[leftmargin=*]  
  \item read manual.txt and show its content  
  \item follow the instructions according to the file content  
  \item report the result to the user  
\end{enumerate} \\ \hline  
\end{tabularx}  
\caption{Two examples of the initial plans and the corresponding final plans.}  
\label{tab:plan}  
\end{table}

After the plan is finalized, the Planner takes action by assigning each sub-task, with phrased queries, to ask the CI to generate a code snippet, leveraging in-domain plugins when necessary. The execution results are sent back to the Planner. Following the ReAct (reasoning and act) design pattern, upon observing the execution results, the Planner may update its plan, request additional information from users, and so on. The process is repeated for subsequent sub-tasks until the entire plan is completed.

When addressing domain-specific scenarios where the LLM's own knowledge is insufficient, the Planner can incorporate external knowledge with ``Examples''. These examples are tailored by scenario developers according to their usage requirements. One example is essentially the chat history between the User, the Planner, and the Code Interpreter, including the plans of the Planner. The Examples could be saved from an online conversation for the purpose of stabilizing the planning process for future requests, or they could be manually prepared. A real planning Example is provided in Appendix \ref{appendix:planner_example}.

\subsection{Code Generator (CG)}

Code Generator is designed to utilize LLMs to automatically synthesize a Python code snippet based on an incoming request. From a high-level view, CG combines the benefits of both the plugin system and code interpreter, allowing the capability of invoking plugins and generating additional code.
Inside the CG, plugins and examples customized by users are leveraged to generate the code.  
\begin{enumerate}[leftmargin=*]
    \item Plugin: In general, plugins can take on various forms such as a web API call, a software module, a customized algorithm, or a deep learning model. Regardless of the form, all variants can be invoked by a function call. Therefore, we have encapsulated the plugins as a Python function within the Code Generator for seamless invocation.
    \item Example: To aid the LLMs in adapting to domain-specific scenarios, examples are designed to help guide LLMs to behave by following the examples. Examples act as a guide for the LLMs to follow and improve their performance by providing contextualized examples. This feature ensures that the generated code aligns with the specific requirements of the user's domain. A real code generation Example is provided in Appendix \ref{appendix:code_generation}.
\end{enumerate}

The CG has the ability to generate code that exclusively calls upon plugins, code that does not use any plugins, or a combination of both. The code that doesn't depend on plugins is specifically tailored to handle requests that cannot be met using only existing plugins.
In the motivating example presented in Section \ref{sec:motivation}, there is a plugin called `anomaly\_detection' that can identify anomalies in data. In this scenario, the CG would first invoke the plugin through a single line of code. Subsequently, the CG would generate multiple lines of additional code to visualize the detected anomalies, as requested by the user.


The knowledge of LLMs is often limited, particularly in domain-specific scenarios. In many cases, there are in-domain tools better suited for completing a specific task. Therefore, the Plugin system can enhance LLMs by enabling them to leverage existing tools. The Plugin system consists of the following two components:

\begin{enumerate}[leftmargin=*]
\item{\textbf{Plugin Schema}} LLMs need to understand the capabilities of a plugin, including its arguments and return values. This information is embedded in the plugin schema file (in YAML format) and could be customized by users. The plugin schema comprises the name, metadata, plugin description, arguments, and return values. For each argument and return value, users should provide its name, type, and description. The plugin schema is then supplied to LLMs as part of the prompt to generate the code.
\item{\textbf{Plugin Implementation}} The plugin implementation defines how each plugin is executed using Python in Code Executor, where the arguments and return values should correspond with the Plugin Schema. Various plugins can be implemented in Python code, such as a domain-specific algorithm, a software module, a deep learning model or a Web API call. It is important to note that LLMs do not need to know the plugin implementation details to generate code. On the contrary, the Plugin Schema is used for code generation.
\end{enumerate}

Appendix \ref{appendix:plugin} shows an example of the plugin schema and implementation.



Even with explicit instructions added to the prompt, such as restricting the call to plugin functions, prohibiting local file system modifications, or disallowing certain package installations, the LLM may still generate code that disregards these instructions. This necessitates a post-verification process to ensure the generated code is safe to execute. Following code generation, the CG parses the code into an Abstract Syntax Tree (AST) and examines it line by line for any violations. If violations are detected, an error message is reported, and a new code snippet is generated. The CG will retry this process several times; if violations persist, the CG will notify the planner of its failure to generate compliant code.


Code auto-correction shares similarities with the post-verification process. If the generated code fails to execute, the exception message is reported to the CG, prompting it to retry generating the code. The primary difference is that post-verification operates within the CG, while code auto-correction relies on the CE to run the code and report errors. Notably, if the execution fails, the Planner can directly redirect the message to the CG instead of synthesizing the report via the LLM. \method permits the code to regenerate up to a maximum of three times.

\subsection{Code Executor (CE)}

The Code Executor (CE) receives the code generated by the CG, collects dependent modules and plugin definitions, executes it, preserves context information such as logs, and returns an execution result to the Planner. 
In \method, we implement CE based on the Python Jupyter kernel. 
To prevent interference between different sessions, CE maintains a separate Jupyter process for each session. In our current implementation, we support two execution modes: local and container.
In the local mode, the Jupyter kernel is launched as a local process, while in the container model, the kernel is running inside a Docker container. When a session concludes, the Jupyter process is terminated. The execution result contains the following parts which are returned to the planner.

\begin{itemize}[leftmargin=*]
    \item \textbf{Return Code}: a successful code execution will return code 0 and otherwise 1.
    \item \textbf{Logs}: Logs are generated in two ways: (1) the stdout/stderr output of the program, and (2) log messages recorded using a logging utility within the plugins. 
    \item \textbf{Output}: This is the output of the Jupyter cell running the generated code.  
    \item \textbf{Artifacts}: The generated code or the plugin calls may produce artifacts such as a CSV file or an image. The user can download the artifacts via the provided URLs. 
\end{itemize}

If the CE fails to execute the code from the CG, it will report the error logs to the CG, who will attempt to revise the code to fix the issue. All the error information and the failed code are kept in the CG's conversation history so that the CG is aware of the full execution history and can precisely understand the state of the CE.

\subsection{Experiences and Personalization}
In practical scenarios, when a user tasks \method with solving a complex problem, the system may initially falter. However, with repeated attempts and additional instructions from the user, \method can eventually find a solution. A challenge arises when a user presents a similar or identical problem at a later time: \method struggles to provide the correct solution promptly because it lacks the capability to remember past experiences. To address this, we have introduced a feature known as \textit{experience memory} within the Memory module of \method.  
   
Through this mechanism, a user can command \method to record the chat history. The system then distills `experience tips' from this history, which encapsulate actionable insights about what to do—or what not to do—in response to requests akin to those encapsulated in the tips. These insights are then stored in an `experience pool'. When \method encounters similar requests in the future, it will draw upon this pool to inform its strategy for planning and code generation. The retrieved experiences are integrated into the prompts used by the Planning and Code Generation (CG) components, thereby enhancing \method's response to subsequent, similar requests.  
   
The utility of the \textit{experience memory} extends beyond merely aiding in the resolution of challenging problems; it also serves to capture and incorporate user preferences into interactions with the agent. For instance, if a user exhibits a predilection for plotting charts with specific color schemes, this preference is gleaned from the chat history and preserved as an experience tip. Consequently, such personalized preferences are automatically considered in relevant future tasks.

\section{Expansion to Multi-Agent Architecture}  
This section elucidates the rationale and methodologies for transitioning \method into a multi-agent system architecture. The shift towards a multi-agent paradigm confers multiple advantages:  
  
\begin{itemize}  
    \item \textbf{Modularity:} It is advantageous to decompose a complex system into a consortium of agents, where each is tasked with a discrete set of functions. This modularity enhances the manageability and maintainability of the system.  
    \item \textbf{Extensibility:} The multi-agent architecture allows for the facile addition of new functionalities. One can introduce new agents with the requisite capabilities into the existing framework without necessitating alterations to the core codebase.  
\end{itemize}  
  
\begin{figure}[ht]
    \centering
    \includegraphics[width=5in]{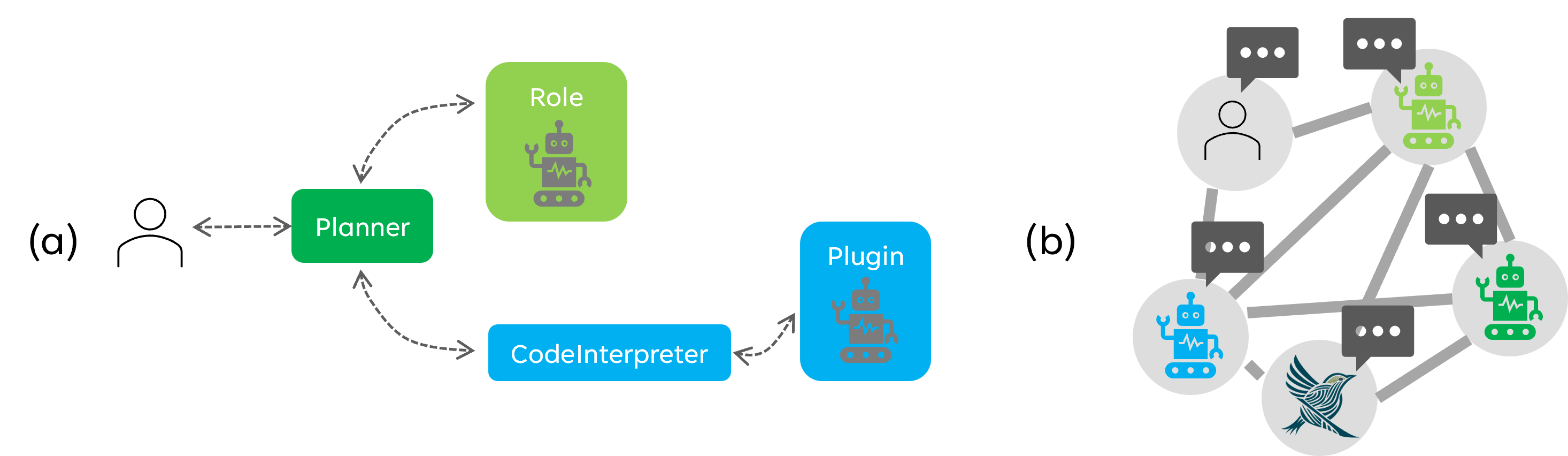}
    \caption{\method in a multi-agent environment.}
    \label{fig:multi_agents}
\end{figure}

There are two principal strategies for incorporating \method within a multi-agent environment, as depicted in Figure \ref{fig:multi_agents}:  
  
\begin{enumerate}  
    \item \textbf{Agent Collaboration via Plugins or Roles:} As shown in Figure \ref{fig:multi_agents}(a), one strategy involves a \method-powered agent invoking other agents through plugins, or encapsulating the functionality of existing agents within newly defined roles. The choice between implementing a plugin or a role is influenced by the specific function provided by the external agent.  
      
    \item \textbf{Integration into an Existing Framework:} The alternative strategy, illustrated in Figure \ref{fig:multi_agents}(b), embeds \method-enhanced agents into a pre-existing multi-agent framework. This integration may require the establishment of a coordination mechanism to govern the interactions among the agents.  
\end{enumerate}

\section{Evaluation}

The evaluation of an LLM-based agent's performance can be challenging. Current evaluation methods often treat the LLM agent as a function that maps input data to output data. When assessing the agent's performance on a multi-step task, the evaluation process resembles a chain of calls to a stateful function. Typically, the agent's output is compared to a ground truth or a reference output to judge its effectiveness. As the agent's output is in natural language, evaluation is commonly conducted by matching keywords or phrases in the output to the ground truth.

However, this evaluation method has limitations due to its rigid nature. It may struggle to effectively evaluate long and complex outputs, especially when matching keywords is not sufficient. Handling different formats, such as dates or numbers, can pose challenges for the evaluation method. Additionally, the method should ideally exhibit a level of understanding similar to that of a human, allowing for contextual comprehension and interpretation of the output. For instance, when different agents are tasked with the same objective, they may exhibit varying behaviors while still producing correct outputs.

The below example in Table \ref{tab:agent_eval_exp} illustrates this point:

\begin{table}[ht]
\centering
\resizebox{0.65\columnwidth}{!}{%
\begin{tabular}{l}
\hline
\begin{tabular}[c]{@{}l@{}}Human: What is the weather today?\\ Agent 1: It is sunny today in New York.\end{tabular}                                                                 \\ \hline
\begin{tabular}[c]{@{}l@{}}Human: What is the weather today?\\ Agent 2: Do you want to know the weather in New York today?\\ Human: Yes.\\ Agent 2: It is sunny today.\end{tabular} \\ \hline
\end{tabular}%
}
\caption{Two agents answering the same question.}
\label{tab:agent_eval_exp}
\end{table}

Compared to Agent 1, Agent 2 asks for confirmation before providing the answer, which requires more interaction with the user. However, both agents provide the correct answer to the question. However, if the evaluation method takes the agent as a function, it may not be able to handle the different behaviors of the agents and consider Agent 2 as incorrect (as the first response does not match the ground truth, e.g., "sunny").

This leads us to propose a more adaptable evaluation approach that introduces two new roles: the Examiner and the Judge. In this approach, for each test case, the Examiner is initially provided with the task description and assumes the responsibility of supervising the conversation with the evaluation target -- the agent. The Examiner has the authority to ask questions to the agent and must ensure that the conversation aligns with the task at hand. Additionally, the evaluation agent is permitted to seek clarification on the task by posing questions to the Examiner. Notably, the Examiner is solely responsible for providing the task description and is prohibited from offering any hints or solutions.

Once the evaluation target presents a solution, the Examiner concludes the conversation and forwards the solution to the Judge for evaluation against the ground truth. This method stands in contrast to the traditional evaluation approach, as it effectively mitigates the limitations previously mentioned.

\subsection{DataSets}\label{subsec:datasets}

\paragraph{Eval-Cases} \method includes a set of test cases specifically designed to verify that it meets our design goals. These test cases cover several aspects, such as plugin usage, code generation, plan decomposition, reasoning and action, stateful execution, security, etc. The test cases also examine common agent skills like web searching and document retrieval. There are a total of 23 test cases. 

\paragraph{DS-1000} DS-1000, as introduced in the work by Lai et al. \cite{Lai2022DS1000}, serves as a code generation benchmark specifically tailored to assess the capability of Language Model Models (LLMs) in generating code for data science-related questions gathered from StackOverflow. Each test case in the benchmark comprises a problem description followed by a sample code snippet that requires completion. The sample code can be of two types: either requiring completion at the end of the snippet or requiring insertion within the existing code. Furthermore, each question is linked to a Python package, such as Pandas, Numpy, Scipy, Tensorflow, or Matplotlib, as an evaluation dependency. Notably, to heighten the evaluation's complexity, the creators of DS-1000 \cite{Lai2022DS1000} intentionally modified the original questions in various ways.

Upon reviewing DS-1000, it becomes evident that its original design did not specifically cater to evaluating an agent's performance. Although agents could potentially respond to the task of completing the missing code, their role would be limited to that of a chatbot. To address this limitation, we have undertaken a transformation of the test cases to render them more suitable for evaluating an agent's performance. Firstly, all test cases have been converted into insertion test cases. Additionally, we have modified the task to involve filling in the blanks within code snippets, executing the code, and subsequently reporting the completed code. To facilitate the code execution requirement, we have filtered out a subset of test cases where the given sample code is not executable. Specifically, test cases with a \textit{problem\_id} greater than 817 have been disregarded, as a majority of cases in this subset contain code snippets that cannot be run \footnote{The sample codes of these cases typically contain expressions like $df=load\_data()$, which represent the data loading function, but the actual implementation of the function is missing.}. The remaining 816 test cases serve as the basis for evaluating the agent's performance. An example test case after transformation is presented in Appendix \ref{appendix:ds-1000}.

\paragraph{InfiAgent-DABench} InfiAgent-DABench, introduced in the work by Hu et al. \cite{hu2024infiagentdabench}, serves as a benchmark designed to assess agents' performance on data analytics tasks. The benchmark comprises a total of 258 test cases, each accompanied by an input file in CSV format. Additionally, each test case presents one or more questions related to the data within the file. In our evaluation of the test cases using the \method, we have slightly adapted each case to involve the initial task of loading a data file, followed by answering the questions based on the loaded data. An example test case after transformation is presented in Appendix \ref{appendix:dabench}.

\paragraph{DSEval} DSEval was proposed by Zhang et al. \cite{zhang2024benchmarking} to evaluate the performance of data science agents. It consists of four benchmarks, namely Exercise, SO, LeetCode, and Kaggle. In total, there are 294 problem sets, where each problem set is a Python file containing multiple problems and each problem has a user input question and the groundtruth code. In evaluation, DSEval will execute both groundtruth code and the code generated by an agent, and then compare the execution output, the Python namespace, etc. We follow the design of DSEval framework with some slight changes to the evaluation code to make it fit for TaskWeaver. The \textit{Pass Rate} (\%PASS) is the number of problems passed divided by all problems in the benchmark, and is used as the metric to assess the performance.

\subsection{Overview of Evaluation Result}

We conducted evaluations on all the test cases from the datasets mentioned earlier. Each test case awards one or several points upon successful completion. As an agent provides solutions, we count the points earned and compute a normalized score ranging from 0 to 1 to represent the test case's final score. We put \method to the test using various LLMs as the underlying model. For each LLM, we performed at least two runs of all test cases and calculated the mean score, which we present as the LLM's overall performance score. The summarized results are shown in the table below.

\begin{table}[th]
    \centering
    \begin{tabular}{l|c|c}
    \toprule
      Benchmarks  & GPT3.5 & GPT4  \\ \hline
        Eval-Cases & 0.42 & 0.87  \\ \hline
        DS-1000 & 0.40 & 0.60 \\ \hline
        InfiAgent-DABench & 0.70 & 0.88 \\ \hline
        DSEval & 0.36 & 0.72 \\
    \bottomrule
    \end{tabular}
    \caption{Evaluation Results on Benchmarks}
    \label{tab:my_label}
\end{table}

The performance of GPT-4 consistently surpasses that of GPT-3.5, underscoring its enhanced capabilities in comprehending problems, strategizing task execution, and generating code. When evaluated across three external benchmark datasets, both GPT-3.5 and GPT-4 achieve their highest scores on the InfiAgent-DABench. This superior performance can be attributed to \method's proficiency in tasks that involve data loading followed by analytical operations. However, when it comes to the DS-1000 dataset, which is derived from StackOverflow queries, the models face challenges. The ambiguity of the questions and the complexity of evaluating answers, particularly those involving DataFrames, make accurate assessment difficult. This suggests that while GPT-4’s advancements are notable, there is still room for improvement in processing and generating precise answers from less structured and clear data sources.


\subsection{Evaluations with the Eval-Cases in TaskWeaver}

The detailed evaluation results for the 23 test cases are depicted in Table \ref{table:combined_results}. Although these cases were crafted to assess compliance with our predefined requirements, it is important to note that their successful completion by \method is not unequivocally assured. This is attributed to various factors, including the inherent variability in LLM performance. A notable observation from our study is that the performance of a given LLM endpoint is not consistent, exhibiting fluctuations over time and across different host machines. Such variability poses a significant challenge for applications based on LLMs.   
  
In Table \ref{table:combined_results}, the `Score' column enumerates the points accrued across various test cases when utilizing GPT3.5 and GPT4. Meanwhile, the `Normalized Score' column provides a proportional representation of these scores, scaled relative to the maximum possible points for each case. On average, the normalized scores for GPT3.5 and GPT4 stand at 0.42 and 0.85, respectively, indicating a substantial performance differential between the two models.  

\begin{table}[htbp]  
\centering  
\begin{tabular}{r|l|r|r|r|r}  
 \toprule
\multirow{2}{*}{\textbf{ID}} & \multirow{2}{*}{\textbf{Case Name}}
 & \multicolumn{2}{c|}{\textbf{Score}} & \multicolumn{2}{c}{\textbf{Normalized Score}} \\ \cline{3-6} 
 & & \textbf{GPT3.5} & \textbf{GPT4} & \textbf{GPT3.5} & \textbf{GPT4} \\ \hline  
1 & web\_search            & 2 & 2   & 1    & 1 \\ \hline  
2 & web\_search\_calc       & 0 & 3   & 0    & 1        \\ \hline  
3 & echo                  & 0 & 2   & 0    & 1        \\ \hline  
4 & sample\_code           & 2 & 2   & 1    & 1        \\ \hline  
5 & shopping\_plan         & 0 & 2.5 & 0    & 0.71 \\ \hline  
6 & save\_file             & 0 & 5   & 0    & 1        \\ \hline  
7 & context\_length        & 0 & 10  & 0    & 0.67 \\ \hline  
8 & rag                   & 1 & 1   & 1    & 1        \\ \hline  
9 & stock\_forecasting     & 3 & 1   & 0.75 & 0.25     \\ \hline  
10 & delete\_files          & 0 & 1   & 0    & 1        \\ \hline  
11& get\_secret\_key        & 1 & 1   & 1    & 1        \\ \hline  
12& planner\_consolidation & 0 & 2   & 0    & 0.67 \\ \hline  
13& stateful              & 1 & 1   & 1    & 1        \\ \hline  
14& anomaly\_detection     & 0 & 4   & 0    & 0.67 \\ \hline  
15& data\_processing       & 0 & 3   & 0    & 1        \\ \hline  
16& list\_files            & 1 & 0   & 1    & 0        \\ \hline  
17& auto\_plugin\_selection & 0 & 2   & 0    & 1        \\ \hline  
18& run\_in\_container      & 1 & 1   & 1    & 1        \\ \hline  
19& file\_chain            & 0 & 1   & 0    & 1        \\ \hline  
20& response\_format       & 0 & 1   & 0    & 1        \\ \hline  
21& command\_line          & 1 & 1   & 1    & 1        \\ \hline  
22& calc\_mean             & 0 & 1   & 0    & 1        \\ \hline  
23& plugin\_only           & 1 & 1   & 1    & 1        \\  \hline
 & Averge & 0.61 & 2.11 & 0.42 & 0.87 \\  
 \bottomrule
\end{tabular}  
\caption{Combined results from GPT-3.5 and GPT-4}  
\label{table:combined_results}  
\end{table}  

Upon detailed examination of Table \ref{table:combined_results}, it becomes evident that GPT3.5 surpasses GPT4 in only two instances, specifically in test cases \#9 and \#16. In test case \#9, which involves predicting stock prices, both GPT3.5 and GPT4 successfully downloaded historical price data. However, only GPT3.5 managed to accurately predict future prices within the evaluation period. For test case \#16, which requires listing files in the current working directory, GPT4 did not execute the task, citing an inability to access the local file system, while GPT3.5 completed the task without issue.

\subsection{Evaluations with the DS-1000 dataset}

\begin{figure}[ht]  
  \centering  
  \begin{minipage}[b]{0.4\linewidth}  
    \includegraphics[width=\linewidth]{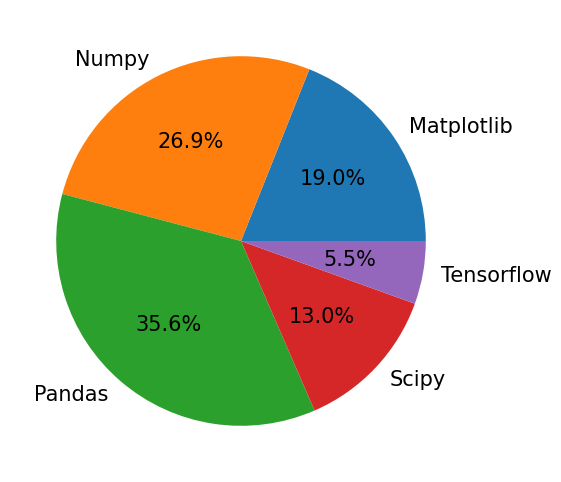}  
    \caption{Percentage of test cases in different libraries.}  
    \label{fig:cnt_lib}  
  \end{minipage}  
  \hfill 
  \begin{minipage}[b]{0.4\linewidth}  
    \includegraphics[width=\linewidth]{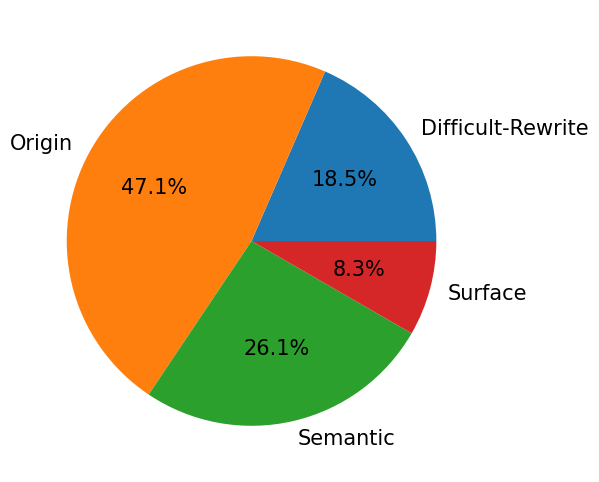}  
    \caption{Percentage of test cases in different perturbation types.} 
    \label{fig:cnt_type}  
  \end{minipage}  
\end{figure}

Following the transformation process in Section \ref{subsec:datasets}, the DS-1000 dataset comprises 816 individual test cases. These cases are categorized across five widely-utilized libraries dedicated to data analytics tasks, with the distribution illustrated in Fig. \ref{fig:cnt_lib}.
To modify the complexity of the original test cases, the developers of the DS-1000 have introduced various perturbations. The variety of these perturbation types and their respective frequencies within the test cases are depicted in Fig. \ref{fig:cnt_type}.
  
\begin{figure}[ht]  
  \centering  
  \begin{minipage}[b]{0.45\linewidth}  
    \includegraphics[width=\linewidth]{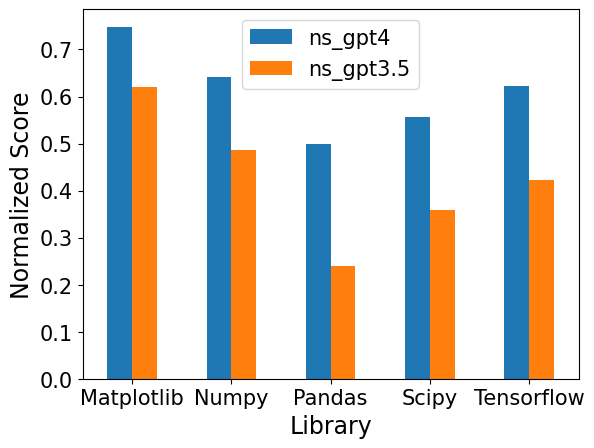}  
    \caption{Normalized scores of test cases in different libraries.}  
    \label{fig:ns_lib}  
  \end{minipage}  
  \hfill 
  \begin{minipage}[b]{0.45\linewidth}  
    \includegraphics[width=\linewidth]{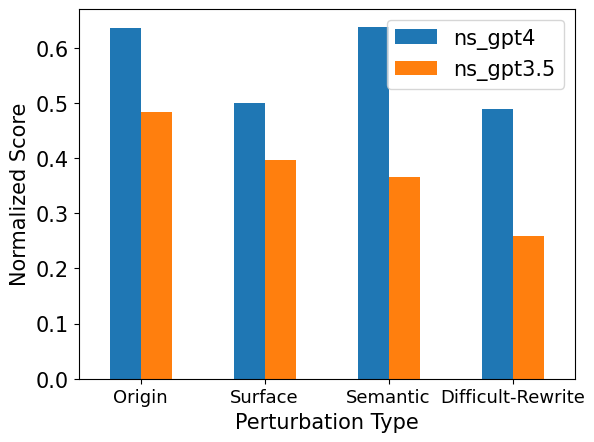}  
    \caption{Normalized scores of test cases in different perturbation types.} 
    \label{fig:ns_type}  
  \end{minipage}  
\end{figure}

Figure \ref{fig:ns_lib} presents the normalized scores for various libraries when processed by two different Large Language Models (LLMs), specifically GPT4 and GPT3.5. On average, GPT4 attains a normalized score of 0.6, surpassing the average score of 0.4 achieved by GPT3.5. An analysis of performance across the five libraries indicates a consistent pattern for both models, with Matplotlib recording the highest normalized score and Pandas the lowest. In light of the lower scores for Pandas, we conducted a closer examination of the cases where the models underperformed. A common issue identified was the challenge in accurately determining the correctness of Pandas test cases, which often require a comparison between two DataFrames. Such comparisons are prone to various errors, including discrepancies in indexes or data types. Nevertheless, these mismatches might not always signify actual failures in a practical context, as they are generally simple to rectify and might not pose a real problem.

Figure \ref{fig:ns_type} displays the normalized scores associated with various perturbation types. It is observable that the performance of GPT3.5 deteriorates as the complexity of the perturbations escalates, which corroborates our initial predictions. Conversely, GPT4's performance does not mirror that of GPT3.5, particularly in the context of Semantic perturbations. This divergence can be attributed to GPT4's enhanced semantic comprehension abilities, which enable it to handle such perturbations more adeptly than GPT3.5, suggesting that these modifications do not pose the same level of difficulty for more advanced models.

\subsection{Evaluations on the InfiAgent-DABench dataset}

\begin{figure}[ht]  
  \centering  
  \begin{minipage}[b]{0.45\linewidth}  
    \includegraphics[width=\linewidth]{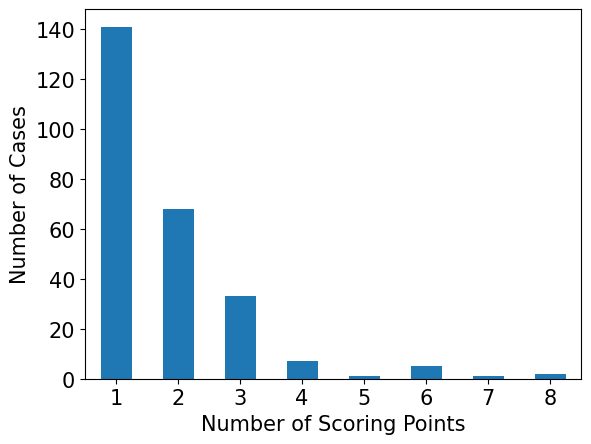}  
    \caption{Number of test cases with different scoring points.}  
    \label{fig:cnt_points}  
  \end{minipage}  
  \hfill 
  \begin{minipage}[b]{0.45\linewidth}  
    \includegraphics[width=\linewidth]{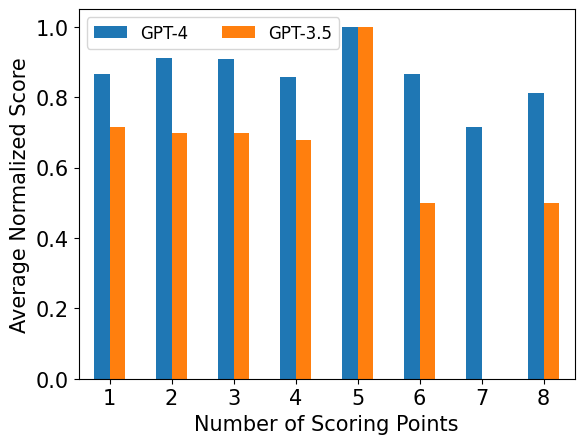}  
    \caption{Normalized scores of test cases with different scoring points.} 
    \label{fig:ns_points}  
  \end{minipage}  
\end{figure}

All test cases in the InfiAgent-BABench are about analyzing the data loaded from a CSV file and then answering one or more questions. Figure \ref{fig:cnt_points} shows the distribution of the number of scoring points of the questions in this dataset. Most of the test cases have less than 3 questions. We conducted the evaluations with the test cases and collected the scoring points gained by the agent, one scoring point for each question. Then, we calculated the normalized score for each test case, i.e., a value between 0 to 1. Figure \ref{fig:ns_points} shows the normalized scores of test cases with different scoring points. The average normalized scores of GPT4 and GPT3.5 are 0.88 and 0.70, respectively. 

\subsection{Evaluations on DSEval dataset}

\begin{table}[h!]  
\centering  
\begin{tabular}{lcccc}  
\toprule  
Dataset
& \%PASS & \%PASS w/o PE & \%PASS w/o IV & \%PASS w/o PE \& IV \\  
\hline  
\multirow{2}{*}{Leetcode} & 42.50 & 60.00 & 42.50 & 60.00 \\  
& 75.00 & 77.50 & 75.00 & 77.50 \\  \hline
\multirow{2}{*}{SO} & 37.13 & 46.04 & 42.08 & 50.99 \\  
& 70.79 & 77.23 & 73.76 & 80.20 \\  \hline
\multirow{2}{*}{Kaggle} & 4.80 & 6.31 & 5.81 & 7.32 \\  
& 35.35 & 41.67 & 44.19 & 50.51 \\  \hline
\multirow{2}{*}{Exercise} & 20.32 & 25.67 & 21.93 & 27.27 \\  
& 63.64 & 75.40 & 66.84 & 78.61 \\  
\bottomrule  
\end{tabular}  
\caption{Comparison of GPT-3.5 and GPT-4 results on DSEval Benchmarks.}  
\label{tab:ds_eval}  
\end{table}  

In this experiment, we conducted a comparative study of GPT-3.5 and GPT-4 across multiple datasets on DSEval benchmark, revealing significant enhancements in GPT-4's performance. The \%PASS denotes the pass rate where TaskWeaver's execution result is exactly the same as the groundtruth one. Comparing the executions of two programs is often very hard. Sometimes, although the agent has fulfilled the user's question, it does not imply that the final results are the same and comparable. Two common errors might lead to this situation, Presentation Errors (PE) and Intact Violation (IV). Presentation Errors manifest when the output is nearly correct but is flawed in terms of formatting or the method of presentation. For instance, an agent might neglect to capitalize a column name as required or mistakenly direct results to the console rather than placing them within the designated cell outputs. Intact Violations occur when the solution closely approaches correctness but fails to maintain the data's original state, often seen when intermediate computations are necessary, and the agent alters the original dataset unnecessarily.

The comparative outcomes, as delineated in Table~\ref{tab:ds_eval}, underscore the substantial strides made by GPT-4 in diminishing the incidence of Presentation Errors (PE) and Intact Violations (IV) across all examined datasets. Notably, within the Leetcode dataset, there was a pronounced increase in the correctness score from 42.50 for GPT-3.5 to 75.00 for GPT-4. Moreover, there was a notable enhancement in the metric representing results unmarred by PE or IV, escalating from 60.00 to 77.50. This pattern of improvement was consistent across additional datasets, including Stack Overflow (SO). While the DSEval benchmark offers a framework for performance evaluation by furnishing ground truth code, it is not without limitations; the rigidity of automated comparisons can lead to undervaluation of the AI's performance, as it may fail to recognize solutions that, from a human perspective, have effectively fulfilled the task requirements. This highlights the need for a more nuanced evaluation approach that can accurately reflect the practical utility of AI-generated code.

\section{Case Studies}
In this section, we demonstrate how to use \method for practical tasks:
\begin{itemize}[leftmargin=*]
    \item Anomaly detection based on data pulled from a database
    \item Stock price forecasting
\end{itemize}

\subsection{Task 1: Anomaly Detection}
We aim to identify anomalies within a time series dataset stored in an SQL database (sqlite3).  
To accomplish this goal, we need to integrate the two plugins for data retrieval and time series anomaly detection:  
   
\begin{itemize}[leftmargin=*]   
    \item \textbf{pull\_data\_sql(nl\_query) $\rightarrow$ sql, dataframe}: Given a natural language query, pull data from a database and return the result in a pandas DataFrame, along with the generated SQL query.  
    \item  \textbf{anomaly\_detection $\rightarrow$ dataframe with anomalies, results description}: This plugin utilizes a straightforward 3-sigma algorithm to identify any abnormal data points in a time series. These data points deviate from the mean value by more than three standard deviations.
\end{itemize}  
   
\textbf{Note}: There is no need to provide plugins for general tasks, such as reading or writing files, as the \method is capable of generating the necessary code for these tasks. This is an advantage of \method that reduces the development effort to bootstrap applications.

Fig.\ref{fig:case_study_task2} in Appendix \ref{sec:appendix_case_study} illustrated the detailed plan generated by the Planner of \method:
\begin{enumerate}[leftmargin=*]
    \item Instruct CodeInterpreter to pull data from the time\_series table in the database.
    \item Confirm the columns to be detected anomalies.
    \item Instruct CodeInterpreter to detect anomalies on the pulled data.
    \item Report the detected anomalies to the user.
\end{enumerate}

In the first step, the Code Interpreter invoked the \textit{pull\_data\_sql} plugin function and automatically filled the required parameters.

\begin{lstlisting}[language=Python, linewidth=\textwidth]
from typing import Tuple
import pandas as pd

query = "SELECT * FROM time_series"
df, description = sql_pull_data(query)
df, description
\end{lstlisting}

After obtaining the time series data from the database, TaskWeaver asked the user to provide the column names as additional information to proceed with anomaly detection.
The user then provided the "ts" and "val" column names to detect anomalies.
With the additional information confirmed, \method proceeded to execute the third step, i.e., detecting anomalies using the \textit{anomaly\_detection} plugin.

\begin{lstlisting}[language=Python, linewidth=\textwidth]
time_col_name = 'ts'
value_col_name = 'val'

anomaly_df, anomaly_description = anomaly_detection(df, time_col_name, value_col_name)
anomaly_df, anomaly_description
\end{lstlisting}

In the end, 11 anomalies were detected and a response was composed for the user.

\subsection{Task 2: Stock Price Forecasting}
The user requested \method to forecast the Nasdaq 100 (QQQ) index prices for the next seven days with the ARIMA model. 
The detailed execution process is shown in Fig.\ref{fig:case_study_task1} in Appendix \ref{sec:appendix_case_study}.
Here is the plan generated by the \method Planner:

\begin{enumerate}[leftmargin=*]
    \item Instruct CodeInterpreter to obtain historical QQQ stock price data.
    \item Preprocess the data.
    \item Train the ARIMA model.
    \item Forecast the stock price for the next 7 days.
    \item Report the predicted stock prices to the user.
\end{enumerate}

After receiving the first subtask from the Planner, the Code Interpreter proceeded to generate a code snippet for the first step. The Code Interpreter thought to itself, "ProgramApe (internal code name) will generate a code snippet to load QQQ historical data using pandas\_datareader library." 
The resulting code snippet is as follows:
\begin{lstlisting}[language=python, linewidth=\textwidth]
import pandas_datareader as pdr
import datetime
start_date = datetime.datetime.now() - datetime.timedelta(days=365)
end_date = datetime.datetime.now()
qqq_data = pdr.get_data_yahoo('QQQ', start=start_date, end=end_date)
qqq_data
\end{lstlisting}

Nonetheless, the execution encountered a failure and the following error information was reported:

\begin{lstlisting}[language=xml, linewidth=\textwidth, numbers=none, basicstyle=\ttfamily\color{gray}]
The execution of the generated python code above has failed
During execution, the following messages were logged:
---------------------------------------------------------------------------
TypeError Traceback (most recent call last)
Cell In[1], line 7
4 start_date = datetime.datetime.now() - datetime.timedelta(days=365)
5 end_date = datetime.datetime.now()
----> 7 qqq_data = pdr.get_data_yahoo('QQQ', start=start_date, end=end_date)
8 qqq_data
File \path\to\python\Lib\site-packages\pandas_datareader\data.py:80, in
get_data_yahoo(*args, **kwargs)
79 def get_data_yahoo(*args, **kwargs):
---> 80 return YahooDailyReader(*args, **kwargs).read()
...
\end{lstlisting}

As a result, the Code Interpreter had to modify the code to resolve this issue.

\begin{lstlisting}[language=python, linewidth=\textwidth]
# Import yfinance or install it if it's not available
try:
    import yfinance as yf
except ImportError:
    !pip install yfinance
    import yfinance as yf

# Load the historical price data for QQQ
qqq_data = yf.download('QQQ')
qqq_data
\end{lstlisting}
In the aforementioned code, we can observe that \method employed an alternative approach to search for stock price data using the yfinance library. 
This time, the historical QQQ stock price data was successfully downloaded and the results were forwarded to the Planner for further instructions.

The Planner received the downloaded stock price data and proceeded with the following steps of preprocessing and ARIMA model training. 
Ultimately, \method utilized the fitted ARIMA model to predict the stock prices for the next 7 days.

\section{Design Verification}

In this section, we use a set of tasks to verify that the \method meets our requirements and design considerations. 

\subsection{Planning} 

\paragraph{Task 1: React} 
ReAct\cite{yao2022react} merges Reasoning and Acting for LLMs, providing a comprehensive paradigm for generating verbal reasoning traces and actions to accomplish a given task. 
By leveraging this approach, the system can conduct real-time reasoning to create, update, and adapt plans while observing the external environments, to incorporate additional information into the reasoning process.

To evaluate the ReAct capabilities of \method, we adopted an approach inspired by the Challenge Task presented in \cite{autoGPT_challenge}. 
This involved generating a collection of $N$ files with the naming convention file$_1$.txt, file$_2$.txt, ..., file$_N$.txt. Each file contained the instruction "read file${i+1}$.txt in the same directory to get the further information." 
The goal was to initiate the \method to begin reading file$_1$.txt and sequentially progress to the subsequent files.
The final file in the sequence contained instructions to present a specific key to the user.

The results of the evaluation are illustrated as following: 

\begin{lstlisting}[language=xml, linewidth=\textwidth, numbers=none, basicstyle=\ttfamily\color{gray}]
Human: read the file_A.txt and follow the instructions in it
......

>>> [PLANNER->CODEINTERPRETER]
Please read the content of the file file_A.txt
......

>>> [CODEINTERPRETER->PLANNER]
The execution of the generated python code above has succeeded
The stdout is:
read file_B.txt in the same directory to get the further information.
......

>>> [PLANNER->CODEINTERPRETER]
Please read the content of the file file_B.txt
......

>>> [CODEINTERPRETER->PLANNER]
The execution of the generated python code above has succeeded
The stdout is:
read file_C.txt in the same directory to get the further information.
......

>>> [PLANNER->CODEINTERPRETER]
Please read the content of the file file_C.txt

>>> [CODEINTERPRETER->PLANNER]
The execution of the generated python code above has succeeded
The result of above Python code after execution is: Show the key to user.
The key is 12345.
......

>>> [PLANNER->USER]
The key is 12345.
\end{lstlisting}

In the depicted process, the Planner component of \method initially broke down the task into two steps and initiated the execution of the first step. This initial phase involved reading the contents of the file\_A.txt. Subsequently, the Code Interpreter received the Planner's request and generated a Python code snippet designed to extract the content from the specified file. The generated code included an instruction to "read file\_B.txt in the same directory to acquire further information."

Upon receiving this instruction, the Planner updated the plan, incorporating the directive to load the next file in the sequence. This iterative process continued until \method successfully navigated through the entire chain of files, culminating in the retrieval of the final message contained in the last file. This concluding message held instructions to present the key to the user, signifying the completion of the task.

\paragraph{Task 2: Plan Decomposition} 
The Planner is capable of breaking down a task into multiple subtasks and delegating them to the Code Interpreter for execution one by one. 
However, given the cost and efficiency of LLM, it is not practical to invoke LLM multiple times, particularly for some simple tasks. 
For this reason, we devised a two-phase planning approach that mandates a review of the generated initial plan, for merging sub-tasks with sequential dependencies (refer to Sec.\ref{subsec:intelli_plan} for more details).

As an illustration, we tasked \method with computing the mean and standard deviation of the 'Count' column in the file "/sample\_data/demo\_data.csv" and subsequently verifying the presence of any values exceeding 3 standard deviations from the mean.
The initial plan generated by the Planner is as follows:

\begin{itemize}[leftmargin=*]
    \item 1. Load the data file
    \item 2. Calculate the mean and std of the 'Count' column <sequentially depends on 1>
    \item 3. Check if there are any values larger than 3 std from the mean <sequentially depends on 2>
    \item 4. Report the result to the user <interactively depends on 3>
\end{itemize}

It is noteworthy that the first and second steps can be combined into a single step since they can be accomplished in a single snippet of code without any interventions. 
As a result, the Planner merged them into one: ``Instruct CodeInterpreter to load the data file and calculate the mean and std of the 'Count' column.'' 
By doing so,  we were able to reduce the number of LLM calls from 6 to 3 for the data loading and the mean/std value calculation.

\subsection{Coding and Execution}
\paragraph{Task 1: Plugin-Only Mode} 
In the \method, we offer a plugin-only mode, which exclusively permits the invocation of plugin functions and forbids the generation of arbitrary code. 
To validate this feature, we activate the plugin-only mode in the configuration file and instruct \method to find out the current date time. Since \method can only call the plugin functions, the Planner refuses to execute the code due to the constraints.

\begin{lstlisting}[language=HTML,linewidth=\textwidth, numbers=none, basicstyle=\ttfamily\color{gray}]
Human: generate and execute python code to get the current time
......
>>> [PLANNER->USER]
I'm sorry, but as a Planner, I do not have the capability to execute Python code. You can run the provided code in your local Python environment to get the current time.
......

\end{lstlisting}

\paragraph{Task 2: Stateful Execution}
The Code Interpreter is stateful, meaning it maintains execution states and variables within the same session. 
For example, we initially asked \method to display the column names of ./sample\_data/demo\_data.csv. 
Subsequently, we instructed \method to execute an irrelevant task, such as ``generate 10 random numbers.'' 
Afterward, we requested the mean value of the ``Count'' column in the previously loaded data.
The \method identified the need to use data from the previous chat round and subsequently delivered an appropriate response.

\paragraph{Task 3: Auto Correction}
The Code Interpreter can make mistakes while generating Python code, resulting in execution failures for the executor. 
To address this, we prompt the Code Interpreter to revise its code based on the reported error information. 
We asked \method to calculate the mean value of sample\_data/demo\_data.csv. 
The Code Interpreter initially generated incorrect code due to the absence of the data schema.
\begin{lstlisting}[language=Python, linewidth=\textwidth]
import pandas as pd

data_file_path = '../../../sample_data/demo_data.csv'
df = pd.read_csv(data_file_path)
mean_value = df.mean()
mean_value
\end{lstlisting}

Upon revising the code, it automatically identifies the columns with numerical data types and calculates their average value successfully.

\begin{lstlisting}[language=Python, linewidth=\textwidth]
import pandas as pd

# Load the data file
data_file_path = '../../../sample_data/demo_data.csv'
df = pd.read_csv(data_file_path)

# Calculate the mean value of the loaded data
mean_value = df.mean(numeric_only=True)
mean_value
\end{lstlisting}

\subsection{Safety} 

\paragraph{Task 1: Preventing File Deletion and Secret Key Leakage}
To ensure the safety and security of the execution environment, we have developed a restricted list to prevent certain sensitive operations, which can be customized by developers. When requesting the \method to delete a file in the system folder or retrieve the secret key from global environment variables, it declines to execute these tasks, as they are deemed high-risk operations.

\section{Related Work}

\paragraph{LLM and Prompt Engineering}
Recent advancements in natural language processing have been driven by large language models (LLMs)~\cite{zhao2023survey} such as GPT~\cite{brown2020language}, GPT-4~\cite{OpenAI2023GPT4TR}, Palm~\cite{anil2023palm}, and Llama~\cite{touvron2023llama}. These models have not only revolutionized the field of natural language processing, but also how humans interact with machines through applications such as ChatGPT. LLMs are pre-trained on a vast amount of text data and then fine-tuned with reinforcement learning from human feedback (RLHF) and Instruction Fine-Tuning (IFT)~\cite{brown2020language} to improve their response quality. To improve the performance of LLMs on reasoning and decision-making tasks, various prompting engineering methods have been proposed, including Chain-of-Thought (CoT)~\cite{wei2022chain}, zero-shot-CoT~\cite{kojima2022large}, and ReAct~\cite{yao2022react}. Some of these approaches have also been applied in \method to enhance its performance.

\paragraph{Agent}
Recently, LLM-based agents have gained increasing attention. The fundamental concept is to utilize LLMs as the core controller to make human-like decisions by observing the environment, planning, and taking actions\cite{agent_blog}. There are generally two types of agent systems: single-agent and multi-agent systems. The single-agent system focuses more on planning, observing, and acting within the single agent's own capability, including AutoGPT\cite{autoGPT} and LangChain Agents\cite{langchain}. Conversely, the latter is more concentrated on leveraging multiple agents to work collaboratively. Typical examples include BabyAGI\cite{babyagi}, MetaGPT\cite{hong2023metagpt}, AutoGen\cite{wu2023autogen}, CAMEL\cite{li2023camel}, and Multi-agent Debate (MAD)\cite{liang2023encouraging}. Our \method is a single-agent framework that focuses on converting user requests into code, even for plugin calls.

\section{Conclusion}
In this paper, we introduced \method, a code-first framework for building LLM-powered autonomous agents that addresses the limitations of existing frameworks in handling rich data structures, incorporating domain knowledge, and offering flexibility. 
\method's standout feature is its ability to convert user requests into executable code while treating user-defined plugins as callable functions. 
This approach enables the seamless integration of plugin execution with custom code execution, catering to the diverse requirements of users and providing a more intuitive user experience.
We presented the design and implementation of \method, highlighting its support for complex data structures, flexible plugin usage, and intelligence task planning. 
We also demonstrated \method's ability to leverage the coding capability of LLMs to implement complex logic and incorporate domain-specific knowledge through examples. 
Furthermore, we discussed the efforts made towards the secure execution of generated code and the provision of an easy-to-use interface for developers.
Through various case studies, we showcased the effectiveness of \method in handling different tasks. 
Overall, \method offers a powerful and flexible solution for building intelligent conversational agents. 
As LLMs continue to evolve and improve, \method can facilitate more advanced and sophisticated applications.

\bibliography{references.bib}  

\begin{thebibliography}{10}

\bibitem{autoGPT}
Autogpt.
\newblock Available at: \url{https://github.com/Significant-Gravitas/AutoGPT}.
\newblock Accessed on [11/22/2023].

\bibitem{autoGPT_challenge}
Autogpt challenge.
\newblock Available at:
  \url{https://github.com/Significant-Gravitas/AutoGPT/blob/master/docs/content/challenges/memory/challenge_a.md}.
\newblock Accessed on [11/22/2023].

\bibitem{babyagi}
Babyagi.
\newblock Available at: \url{https://github.com/yoheinakajima/babyagi}.
\newblock Accessed on [11/22/2023].

\bibitem{jarvis}
Jarvis.
\newblock Available at: \url{https://github.com/microsoft/JARVIS}.
\newblock Accessed on [11/22/2023].

\bibitem{langchain}
Langchain.
\newblock Available at: \url{https://www.langchain.com/}.
\newblock Accessed on [11/22/2023].

\bibitem{agent_blog}
Llm powered autonomous agents.
\newblock Available at:
  \url{https://lilianweng.github.io/posts/2023-06-23-agent/}.
\newblock Accessed on [11/22/2023].

\bibitem{openinterpreter}
Openinterpreter.
\newblock Available at:
  \url{https://github.com/OpenInterpreter/open-interpreter}.
\newblock Accessed on [05/08/2024].

\bibitem{semantickernel}
Semantic kernel.
\newblock Available at: \url{https://github.com/microsoft/semantic-kernel}.
\newblock Accessed on [11/22/2023].

\bibitem{transformeragents}
Transformers agents.
\newblock Available at:
  \url{https://huggingface.co/docs/transformers/transformers_agents}.
\newblock Accessed on [11/22/2023].

\bibitem{anil2023palm}
Rohan Anil, Andrew~M Dai, Orhan Firat, Melvin Johnson, Dmitry Lepikhin,
  Alexandre Passos, Siamak Shakeri, Emanuel Taropa, Paige Bailey, Zhifeng Chen,
  et~al.
\newblock Palm 2 technical report.
\newblock {\em arXiv preprint arXiv:2305.10403}, 2023.

\bibitem{claude}
Amanda Askell, Yuntao Bai, Anna Chen, Dawn Drain, Deep Ganguli, Tom Henighan,
  Andy Jones, Nicholas Joseph, Benjamin Mann, Nova DasSarma, Nelson Elhage, Zac
  Hatfield{-}Dodds, Danny Hernandez, Jackson Kernion, Kamal Ndousse, Catherine
  Olsson, Dario Amodei, Tom~B. Brown, Jack Clark, Sam McCandlish, Chris Olah,
  and Jared Kaplan.
\newblock A general language assistant as a laboratory for alignment.
\newblock {\em CoRR}, abs/2112.00861, 2021.

\bibitem{brown2020language}
Tom Brown, Benjamin Mann, Nick Ryder, Melanie Subbiah, Jared~D Kaplan, Prafulla
  Dhariwal, Arvind Neelakantan, Pranav Shyam, Girish Sastry, Amanda Askell,
  et~al.
\newblock Language models are few-shot learners.
\newblock {\em Advances in neural information processing systems},
  33:1877--1901, 2020.

\bibitem{hong2024data}
Sirui Hong, Yizhang Lin, Bang Liu, Bangbang Liu, Binhao Wu, Danyang Li, Jiaqi
  Chen, Jiayi Zhang, Jinlin Wang, Li~Zhang, Lingyao Zhang, Min Yang, Mingchen
  Zhuge, Taicheng Guo, Tuo Zhou, Wei Tao, Wenyi Wang, Xiangru Tang, Xiangtao
  Lu, Xiawu Zheng, Xinbing Liang, Yaying Fei, Yuheng Cheng, Zongze Xu, and
  Chenglin Wu.
\newblock Data interpreter: An llm agent for data science, 2024.

\bibitem{hong2023metagpt}
Sirui Hong, Xiawu Zheng, Jonathan Chen, Yuheng Cheng, Ceyao Zhang, Zili Wang,
  Steven Ka~Shing Yau, Zijuan Lin, Liyang Zhou, Chenyu Ran, et~al.
\newblock Metagpt: Meta programming for multi-agent collaborative framework.
\newblock {\em arXiv preprint arXiv:2308.00352}, 2023.

\bibitem{hu2024infiagentdabench}
Xueyu Hu, Ziyu Zhao, Shuang Wei, Ziwei Chai, Qianli Ma, Guoyin Wang, Xuwu Wang,
  Jing Su, Jingjing Xu, Ming Zhu, Yao Cheng, Jianbo Yuan, Jiwei Li, Kun Kuang,
  Yang Yang, Hongxia Yang, and Fei Wu.
\newblock Infiagent-dabench: Evaluating agents on data analysis tasks, 2024.

\bibitem{jiang2024mixtral}
Albert~Q. Jiang, Alexandre Sablayrolles, Antoine Roux, Arthur Mensch, Blanche
  Savary, Chris Bamford, Devendra~Singh Chaplot, Diego de~las Casas, Emma~Bou
  Hanna, Florian Bressand, Gianna Lengyel, Guillaume Bour, Guillaume Lample,
  Lélio~Renard Lavaud, Lucile Saulnier, Marie-Anne Lachaux, Pierre Stock,
  Sandeep Subramanian, Sophia Yang, Szymon Antoniak, Teven~Le Scao, Théophile
  Gervet, Thibaut Lavril, Thomas Wang, Timothée Lacroix, and William~El Sayed.
\newblock Mixtral of experts, 2024.

\bibitem{kojima2022large}
Takeshi Kojima, Shixiang~Shane Gu, Machel Reid, Yutaka Matsuo, and Yusuke
  Iwasawa.
\newblock Large language models are zero-shot reasoners.
\newblock {\em Advances in neural information processing systems},
  35:22199--22213, 2022.

\bibitem{Lai2022DS1000}
Yuhang Lai, Chengxi Li, Yiming Wang, Tianyi Zhang, Ruiqi Zhong, Luke
  Zettlemoyer, Wen-Tau Yih, Daniel Fried, Sida Wang, and Tao Yu.
\newblock Ds-1000: A natural and reliable benchmark for data science code
  generation.
\newblock {\em ArXiv}, abs/2211.11501, 2022.

\bibitem{li2023camel}
Guohao Li, Hasan Abed Al~Kader Hammoud, Hani Itani, Dmitrii Khizbullin, and
  Bernard Ghanem.
\newblock Camel: Communicative agents for" mind" exploration of large scale
  language model society.
\newblock {\em arXiv preprint arXiv:2303.17760}, 2023.

\bibitem{liang2023encouraging}
Tian Liang, Zhiwei He, Wenxiang Jiao, Xing Wang, Yan Wang, Rui Wang, Yujiu
  Yang, Zhaopeng Tu, and Shuming Shi.
\newblock Encouraging divergent thinking in large language models through
  multi-agent debate.
\newblock {\em arXiv preprint arXiv:2305.19118}, 2023.

\bibitem{OpenAI2023GPT4TR}
OpenAI.
\newblock Gpt-4 technical report.
\newblock {\em ArXiv}, abs/2303.08774, 2023.

\bibitem{radford2018improving}
Alec Radford, Karthik Narasimhan, Tim Salimans, and Ilya Sutskever.
\newblock Improving language understanding with unsupervised learning.
\newblock {\em OpenAI Blog}, 2018.

\bibitem{weihao2024cradle}
Weihao Tan, Ziluo Ding, Wentao Zhang, Boyu Li, Bohan Zhou, Junpeng Yue,
  Haochong Xia, Jiechuan Jiang, Longtao Zheng, Xinrun Xu, Yifei Bi, Pengjie Gu,
  Xinrun Wang, Börje~F. Karlsson, Bo~An, and Zongqing Lu.
\newblock {Towards General Computer Control: A Multimodal Agent For Red Dead
  Redemption II As A Case Study}.
\newblock {\em arXiv preprint arXiv:2403.03186}, 2024.

\bibitem{geminiteam2024gemini}
Gemini Team.
\newblock Gemini: A family of highly capable multimodal models, 2024.

\bibitem{xagent2023}
XAgent Team.
\newblock Xagent: An autonomous agent for complex task solving, 2023.

\bibitem{touvron2023llama}
Hugo Touvron, Thibaut Lavril, Gautier Izacard, Xavier Martinet, Marie-Anne
  Lachaux, Timoth{\'e}e Lacroix, Baptiste Rozi{\`e}re, Naman Goyal, Eric
  Hambro, Faisal Azhar, et~al.
\newblock Llama: Open and efficient foundation language models.
\newblock {\em arXiv preprint arXiv:2302.13971}, 2023.

\bibitem{wang2023survey}
Lei Wang, Chen Ma, Xueyang Feng, Zeyu Zhang, Hao Yang, Jingsen Zhang, Zhiyuan
  Chen, Jiakai Tang, Xu~Chen, Yankai Lin, et~al.
\newblock A survey on large language model based autonomous agents.
\newblock {\em arXiv preprint arXiv:2308.11432}, 2023.

\bibitem{wei2022chain}
Jason Wei, Xuezhi Wang, Dale Schuurmans, Maarten Bosma, Fei Xia, Ed~Chi, Quoc~V
  Le, Denny Zhou, et~al.
\newblock Chain-of-thought prompting elicits reasoning in large language
  models.
\newblock {\em Advances in Neural Information Processing Systems},
  35:24824--24837, 2022.

\bibitem{wu2023autogen}
Qingyun Wu, Gagan Bansal, Jieyu Zhang, Yiran Wu, Shaokun Zhang, Erkang Zhu,
  Beibin Li, Li~Jiang, Xiaoyun Zhang, and Chi Wang.
\newblock Autogen: Enabling next-gen llm applications via multi-agent
  conversation framework.
\newblock 2023.

\bibitem{xi2023rise}
Zhiheng Xi, Wenxiang Chen, Xin Guo, Wei He, Yiwen Ding, Boyang Hong, Ming
  Zhang, Junzhe Wang, Senjie Jin, Enyu Zhou, et~al.
\newblock The rise and potential of large language model based agents: A
  survey.
\newblock {\em arXiv preprint arXiv:2309.07864}, 2023.

\bibitem{yao2022react}
Shunyu Yao, Jeffrey Zhao, Dian Yu, Nan Du, Izhak Shafran, Karthik Narasimhan,
  and Yuan Cao.
\newblock React: Synergizing reasoning and acting in language models.
\newblock {\em arXiv preprint arXiv:2210.03629}, 2022.

\bibitem{ufo}
Chaoyun Zhang, Liqun Li, Shilin He, Xu~Zhang, Bo~Qiao, Si~Qin, Minghua Ma,
  Yu~Kang, Qingwei Lin, Saravan Rajmohan, Dongmei Zhang, and Qi~Zhang.
\newblock {UFO: A UI-Focused Agent for Windows OS Interaction}.
\newblock {\em arXiv preprint arXiv:2402.07939}, 2024.

\bibitem{zhang2024benchmarking}
Yuge Zhang, Qiyang Jiang, Xingyu Han, Nan Chen, Yuqing Yang, and Kan Ren.
\newblock Benchmarking data science agents, 2024.

\bibitem{zhao2023survey}
Wayne~Xin Zhao, Kun Zhou, Junyi Li, Tianyi Tang, Xiaolei Wang, Yupeng Hou,
  Yingqian Min, Beichen Zhang, Junjie Zhang, Zican Dong, et~al.
\newblock A survey of large language models.
\newblock {\em arXiv preprint arXiv:2303.18223}, 2023.

\bibitem{zhou2023agents}
Wangchunshu Zhou, Yuchen~Eleanor Jiang, Long Li, Jialong Wu, Tiannan Wang, Shi
  Qiu, Jintian Zhang, Jing Chen, Ruipu Wu, Shuai Wang, et~al.
\newblock Agents: An open-source framework for autonomous language agents.
\newblock {\em arXiv preprint arXiv:2309.07870}, 2023.

\end{thebibliography}
\bibliographystyle{plain}

\appendix

\section{Case Study Results}
\label{sec:appendix_case_study}

\begin{figure}[!h]
\caption{Case Study - Task 1: Anomaly Detection}
\includegraphics[width=0.95\textwidth]{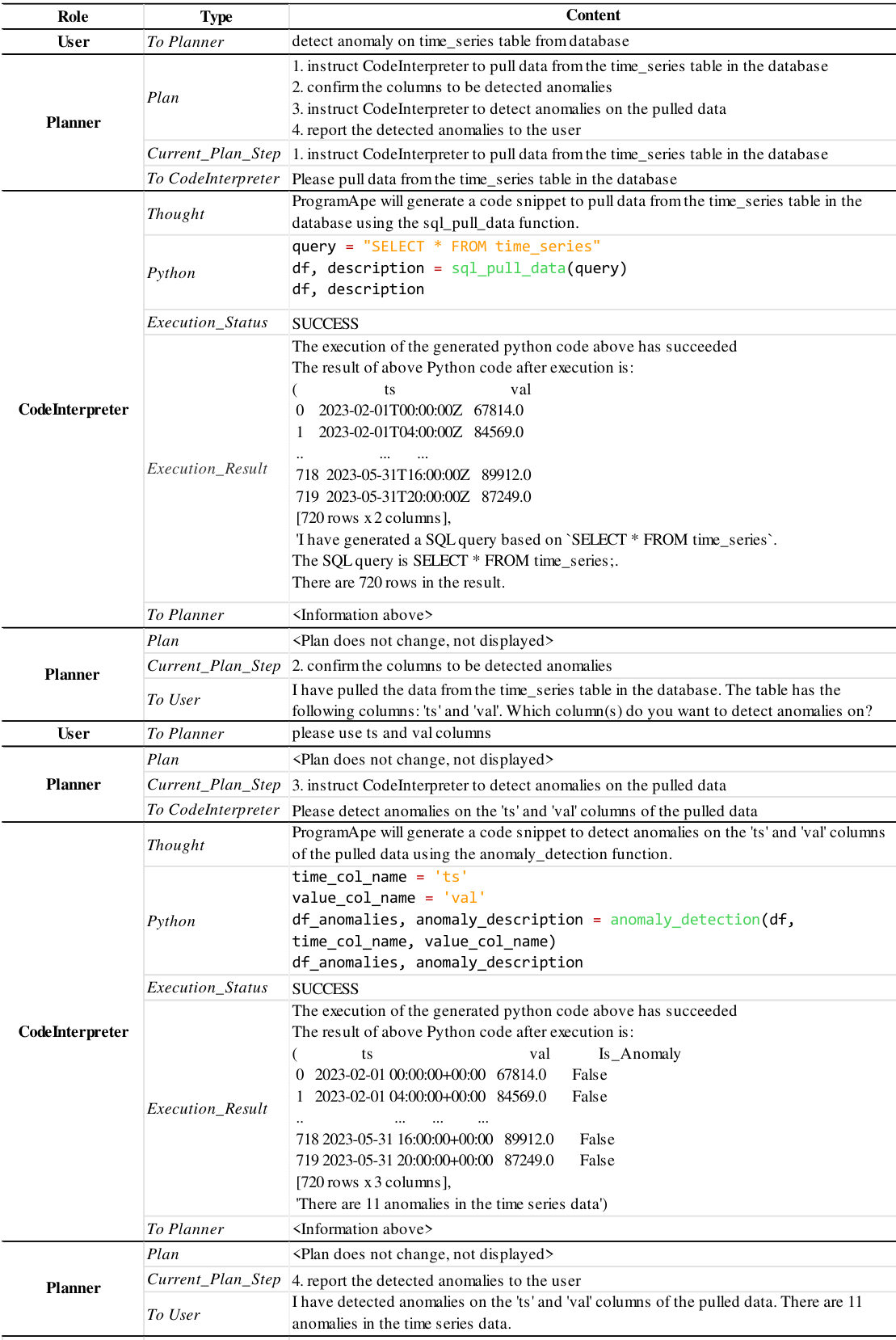}
\label{fig:case_study_task2}
\end{figure}

\begin{figure}[!h]
\centering
 \caption{Case Study - Task 2: Stock Price Forecasting}
\label{fig:case_study_task1}
\includegraphics[width=0.95\textwidth]{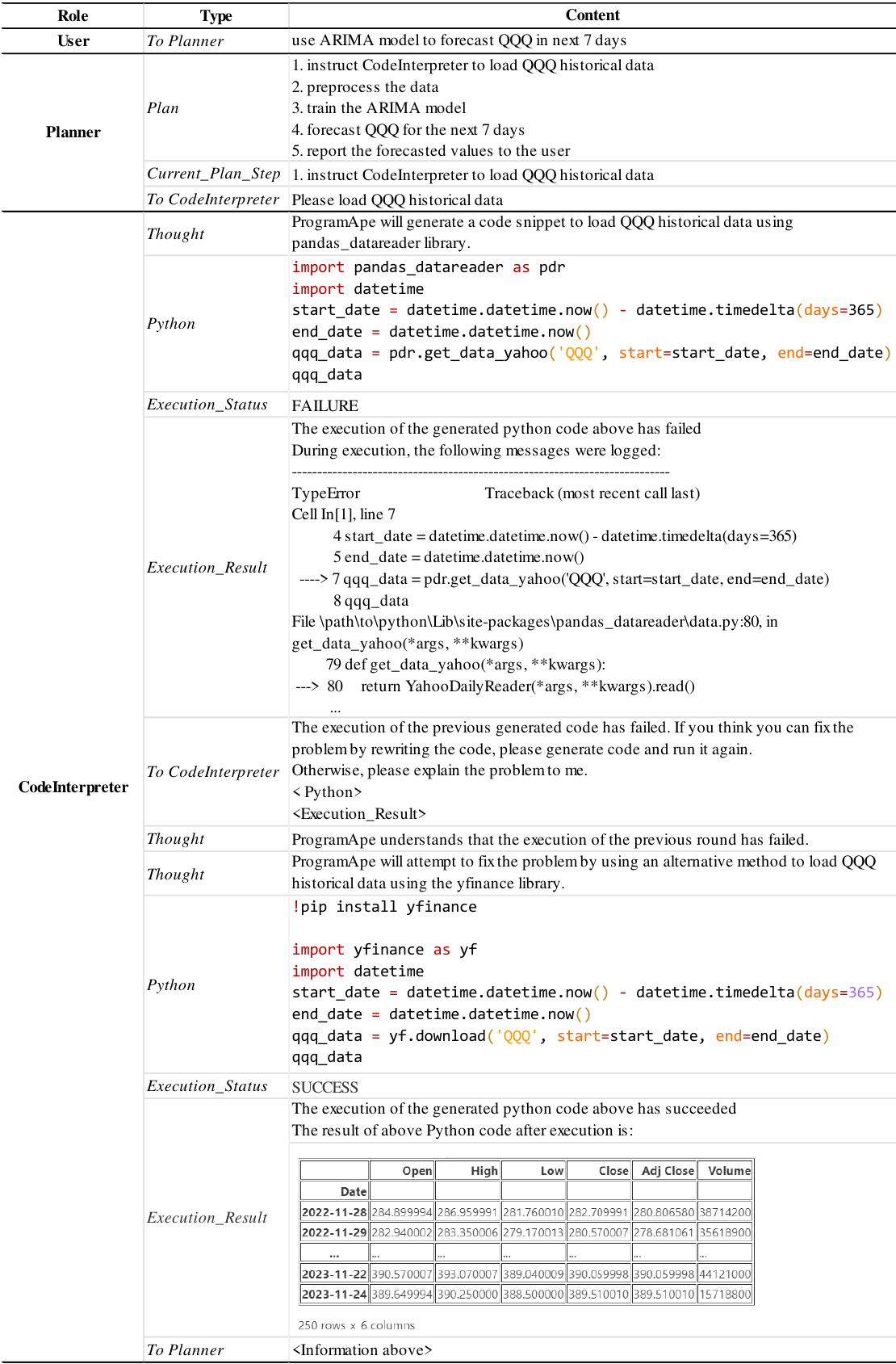}

\end{figure}
\begin{figure}[!htp]
\centering
\includegraphics[width=0.95\textwidth]{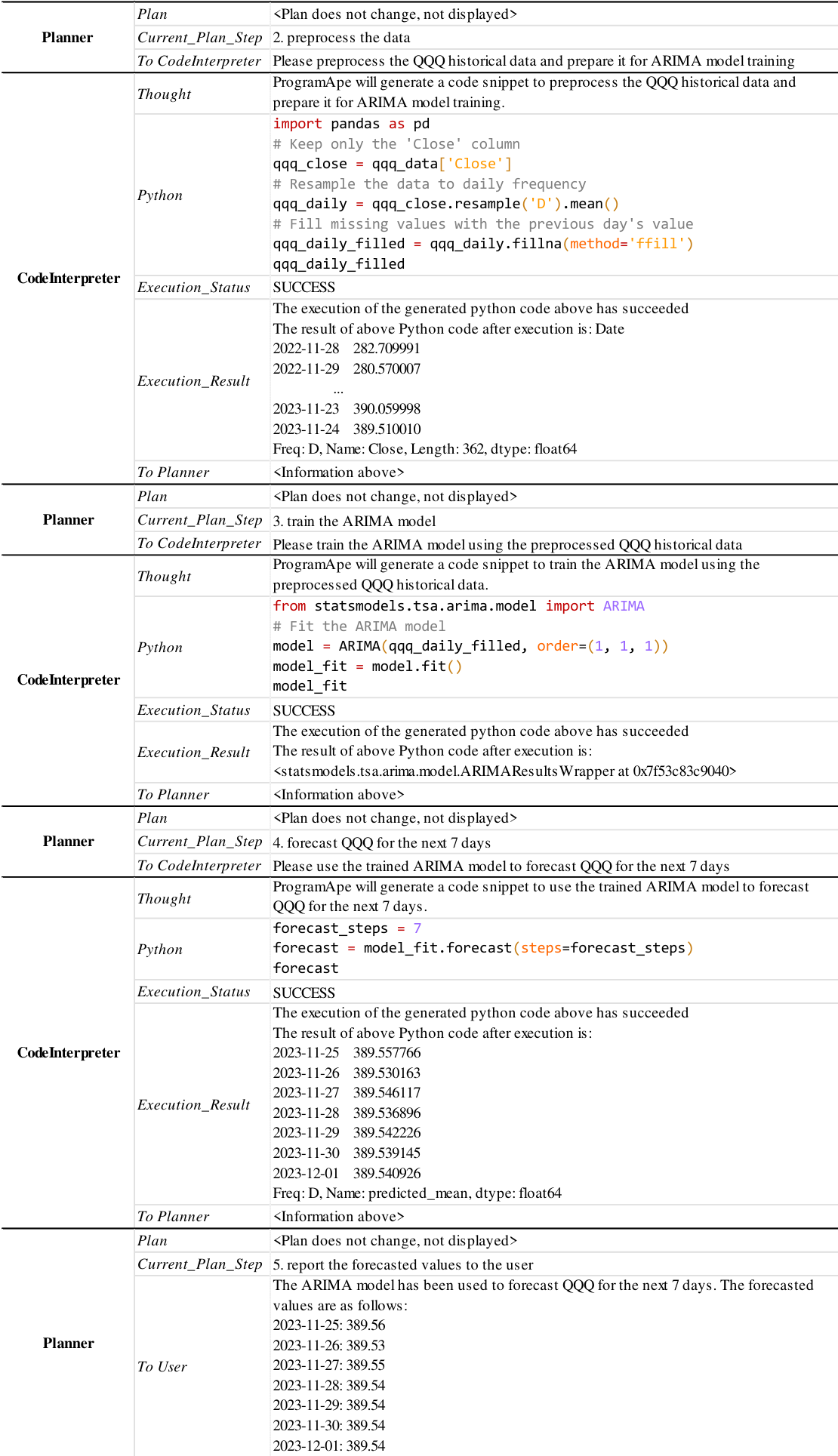}
\end{figure}

\section{Planning Example} \label{appendix:planner_example}
The following is a planning example for the user query of counting the rows of a CSV file. The example consists of 4 posts:
\begin{enumerate}[leftmargin=*]
    \item This is the post from the user containing the user query.
    \item This is the post from the Planner to the Code Interpreter. The post contains 3 attachments that are the initial plan, the plan (final), and the current step in the plan. 
    \item This is the post from the Code Interpreter to the Planner on the execution result.
    \item This is the post from the Planner to the User, i.e., the response after successfully conducting the query.
\end{enumerate}
These 4 posts consist of a round of conversation between the Planner and the User. 

\begin{lstlisting}[language=YAML, linewidth=\textwidth]
user_query: count the rows of /home/data.csv
post_list:
- message: count the rows of /home/data.csv
  send_from: User
  send_to: Planner
  attachment_list:
- message: Please load /home/data.csv and count the rows
  send_from: Planner
  send_to: CodeInterpreter
  attachment_list:
  - type: init_plan
    content: |-
      1. load the data file
      2. count the rows of the loaded data <sequential depend on 1>
      3. report the result to the user <interactive depend on 2>
  - type: plan
    content: |-
      1. instruct CI to load the file and count the rows 
      2. report the result to the user
  - type: current_plan_step
    content: 1. instruct CI to load the file and count the rows 
- message: Load successfully and there are 100 rows
  send_from: CodeInterpreter
  send_to: Planner
  attachment_list:
- message: The file is loaded and there are 100 rows
  send_from: Planner
  send_to: User
  attachment_list:
    - type: init_plan
      content: |-
        1. load the data file
        2. count the rows of the loaded data <sequential depend on 1>
        3. report the result to the user <interactive depend on 2>
    - type: plan
      content: |-
        1. instruct CI to load the file and count the rows 
        2. report the result to the user
    - type: current_plan_step
      content: 2. report the result to the user
\end{lstlisting}

\section{Code Generation Example} \label{appendix:code_generation}
The following is a code generation example for the user query of generating 10 random numbers. This example contains 2 posts:
\begin{enumerate}[leftmargin=*]
\item This is the post from the user containing the user query.
\item This is the post from the Code Interpreter to the Planner including thoughts, the generated code, and the verification and execution results.
\end{enumerate}

\begin{lstlisting}[language=YAML, linewidth=\textwidth]
user_query: generate 10 random numbers
post_list:
- message: generate 10 random numbers
  send_from: Planner
  send_to: CodeInterpreter
  attachment_list: []
- message: The random numbers are 0.2, 0.4, 0.6, ...
  send_from: CodeInterpreter
  send_to: Planner
  attachment_list:
    - type: thought
      content: CI will generate 10 random numbers using np.random.
    - type: python
      content: |-
        import numpy as np
        random_numbers = np.random.rand(10)
        random_numbers
    - type: verification
      content: CORRECT
    - type: code_error
      content: No error is detected.
    - type: execution_status
      content: SUCCESS
    - type: execution_result
      content: The random numbers are 0.2, 0.4, 0.6, ...
\end{lstlisting}

\section{Plugin Example}\label{appendix:plugin}
The following is an example of the specification of a plugin named `anomaly\_detection'. It has 3 input arguments (df, ts\_col, and val\_col), and 2 return values (df and description). 

\begin{lstlisting}[language=YAML, linewidth=\textwidth]
name: anomaly_detection
description: >-
  anomaly_detection function identifies anomalies from an input 
  DataFrame of time series. It will add a new column "Is_Anomaly", 
  where each entry will be marked with "True" 
  if the value is an anomaly or "False" otherwise.

parameters:
  - name: df
    type: DataFrame
    required: true
    description: >-
      the input data from which we can identify the anomalies 
      with the 3-sigma algorithm.
  - name: ts_col
    type: str
    required: true
    description: name of the column that contains the datetime
  - name: val_col
    type: str
    required: true
    description: name of the column that contains the numeric values.

returns:
  - name: df
    type: DataFrame
    description: >-
      This DataFrame extends the input DataFrame with a newly-added 
      column "Is_Anomaly" containing the anomaly detection result.
  - name: description
    type: str
    description: a string describing the anomaly detection results.
\end{lstlisting}

The following shows the Python implementation of the anomaly\_detection plugin. 
\begin{lstlisting}[language=PYTHON, linewidth=\textwidth]
def __call__(self, df: pd.DataFrame, ts_col: str, val_col: str):
    try:
        df[ts_col] = pd.to_datetime(df[ts_col])
    except Exception:
        print("Time column is not datetime")
        return
    
    if not is_numeric_dtype(df[val_col]):
        try:
            df[val_col] = df[val_col].astype(float)
        except ValueError:
            print("Value column is not numeric")
            return
    
    mean, std = df[val_col].mean(), df[val_col].std()
    cutoff = std * 3
    l, u = mean - cutoff, mean + cutoff
    df["Is_Anomaly"] = df[val_col].apply(lambda x: x < l or x > u)
    anomaly_count = df["Is_Anomaly"].sum()
    desc = f"There are {anomaly_count} anomalies in the data"
    
    return df, desc
\end{lstlisting}

\section{DS-1000 Test Case Example}\label{appendix:ds-1000}

The following list shows a transformed test case from the DS-1000 dataset. We put the original problem description inside the <TASK DESCRIPTION> block and added a static header to the test case that explains the task of completing the sample code, running it, and finally presenting the code back. 

\begin{lstlisting}[frame=single, breaklines=true, breakindent=0pt]
The task is to complete the sample code described in the <TASK DESCRIPTION> block below. Complete the code, run it successfully, and finally present the code back. Please "copy and paste" the following task description in your request to ensure that the task description is correct and complete.

<TASK DESCRIPTION>
# Problem
I have the following DataFrame:
    Col1  Col2  Col3  Type
0      1     2     3     1
1      4     5     6     1
2      7     8     9     2
3    10    11    12     2
4    13    14    15     3
5    16    17    18     3


The DataFrame is read from a CSV file. All rows which have Type 1 are on top, followed by the rows with Type 2, followed by the rows with Type 3, etc.
I would like to shuffle the order of the DataFrame rows according to a list. 
For example, give a list [2, 4, 0, 3, 1, 5] and desired result should be:
    Col1  Col2  Col3  Type
2      7     8     9     2
4     13    14    15     3
0     1     2     3     1
3    10    11    12     2
1     4     5     6     1
5    16    17    18     3
...


How can I achieve this?


# Solution
The following is the solution code to the problem statement provided above.
You must complete the code by filling in the missing parts between `### SOLUTION START` and `### SOLUTION END`.
You must keep any code outside of `### SOLUTION START` and `### SOLUTION END` untouched.
Once you have completed the code, run it to check if your solution is correct.
Make sure you keep `### SOLUTION START` and `### SOLUTION END` along with your solution code.


```python

import pandas as pd
import numpy as np


df = pd.DataFrame({'Col1': [1, 4, 7, 10, 13, 16],
                   'Col2': [2, 5, 8, 11, 14, 17],
                   'Col3': [3, 6, 9, 12, 15, 18],
                   'Type': [1, 1, 2, 2, 3, 3]})
List = np.random.permutation(len(df))
### SOLUTION START
result = ... # put solution in this variable
### SOLUTION END
```

</TASK DESCRIPTION>

\end{lstlisting}

\section{InfiAgent-DABench Test Case Example}\label{appendix:dabench}

The following list shows an example of the transformed test case of the InfiAgent-DABench dataset. Each test case involves loading a CSV file and then asking one or more questions concerning the data in the file. The transformation is quite straightforward, only adding the Task section to the original description of the problem. 

\begin{lstlisting}[frame=single, breaklines=true, breakindent=0pt]
# Task
Load the file test_ave.csv and answer the following questions.

# Question
Calculate the mean fare paid by the passengers.

# Constraints
Calculate the mean fare using Python built-in statistics module or appropriate statistical method in pandas. Rounding off the answer to two decimal places.

# Format
@mean_fare[mean_fare_value] where "mean_fare_value" is a floating-point number rounded to two decimal places.
\end{lstlisting}
\end{document}